\documentclass[10pt,twocolumn,letterpaper]{article}

\usepackage{iccv}
\usepackage{times}
\usepackage{epsfig}
\usepackage{graphicx}
\usepackage{amsmath}
\usepackage{amssymb}
\usepackage{multirow}
\usepackage{multicol}
\usepackage{subcaption}
\usepackage{booktabs}
\usepackage{caption}
\usepackage[accsupp]{axessibility}
\usepackage{algorithm}
\usepackage{algorithmic}

\usepackage[breaklinks=true,bookmarks=false]{hyperref}

\makeatletter
\newcommand{\specificthanks}[1]{\@fnsymbol{#1}}

\iccvfinalcopy % *** Uncomment this line for the final submission

\def\iccvPaperID{5458} % *** Enter the ICCV Paper ID here

\ificcvfinal\fi

\begin{document}

%%%%%%%%% TITLE
\title{Revisiting Vision Transformer from the View of Path Ensemble}

\author{Shuning Chang\textsuperscript{\rm 1}\thanks{Work done during an internship at Alibaba Group.}\quad Pichao Wang\textsuperscript{\rm 2}\thanks{Work done at Alibaba Group, and now affiliated with Amazon.}\textsuperscript{\,  \specificthanks{3}}\quad Hao Luo\textsuperscript{\rm 2}\quad Fan Wang\textsuperscript{\rm 2}\quad Mike Zheng Shou\textsuperscript{\rm 1}\thanks{Equal corresponding authors.} \\ 
\textsuperscript{\rm 1}Show Lab, National University of Singapore \quad \textsuperscript{\rm 2}Alibaba Group \\
\small{changshuning@u.nus.edu,\{michuan.lh, fan.w\}@alibaba-inc.com, \{pichaowang, mike.zheng.shou\}@gmail.com}
}

\maketitle
\ificcvfinal\thispagestyle{empty}\fi

%%%%%%%%% ABSTRACT
\begin{abstract}
Vision Transformers (ViTs) are normally regarded as a stack of transformer layers. In this work, we propose a novel view of ViTs showing that they can be seen as ensemble networks containing multiple parallel paths with different lengths. Specifically, we equivalently transform the traditional cascade of multi-head self-attention (MSA) and feed-forward network (FFN) into three parallel paths in each transformer layer. Then, we utilize the identity connection in our new transformer form and further transform the ViT into an explicit multi-path ensemble network. From the new perspective, these paths perform two functions: the first is to provide the feature for the classifier directly, and the second is to provide the lower-level feature representation for subsequent longer paths. We investigate the influence of each path for the final prediction and discover that some paths even pull down the performance. Therefore, we propose the path pruning and EnsembleScale skills for improvement, which cut out the underperforming paths and re-weight the ensemble components, respectively, to optimize the path combination and make the short paths focus on providing high-quality representation for subsequent paths. We also demonstrate that our path combination strategies can help ViTs go deeper and act as high-pass filters to filter out partial low-frequency signals. To further enhance the representation of paths served for subsequent  paths, self-distillation is applied to transfer knowledge from the long paths to the short paths. This work calls for more future research to explain and design ViTs from new perspectives.
\end{abstract}

%%%%%%%%% BODY TEXT
\vspace{-5mm}
\section{Introduction}
%Much of current enthusiasm in application of Transformers~\cite{vaswani2017attention} to vision tasks commences with the Vision Transformer (ViT)~\cite{dosovitskiy2020vit} which can be regarded as alternating layers of Multi-Head Self-Attention (MHSA) and Feed-Forward Network (FFN).
Vision Transformer (ViT)~\cite{dosovitskiy2020vit} consists of alternating layers of Multi-Head Self-Attention (MHSA) and Feed-Forward Network (FFN).
Most follow-ups~\cite{touvron2021training,Xu_2021_ICCV,wang2021pyramid,wu2021cvt,jiang2021all,yu2022metaformer,liu2021swin} focus on polishing these two core modules and create various ViT variants. However, most of them do not break the basic ViT structure, \ie, a stack of transformers containing residual sub-layers, for analysis. 

\begin{figure}[t]
\setlength{\abovecaptionskip}{0.1cm}
    \begin{subfigure}{.17\textwidth}
        \centering
        \includegraphics[width=0.9in]{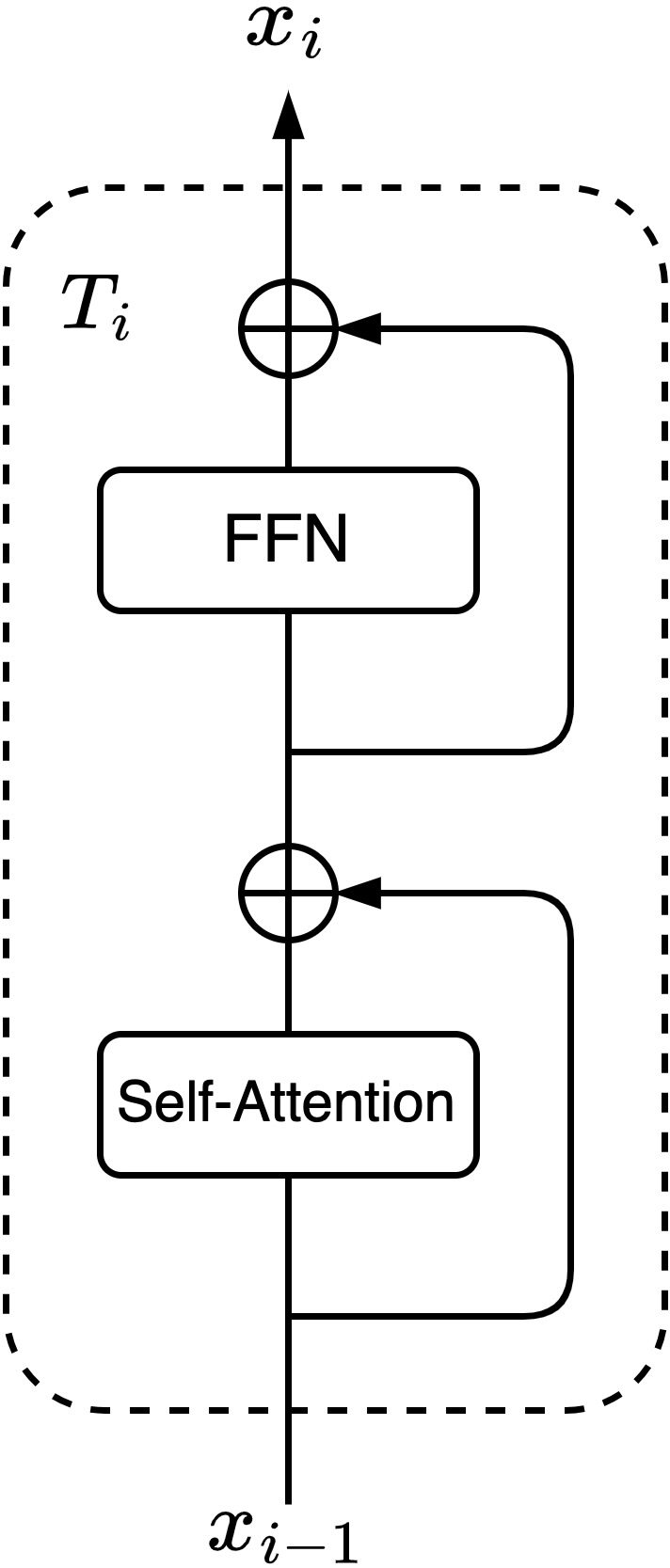}
        \caption{}
        \label{figure1a}
    \end{subfigure}
    \begin{subfigure}{.33\textwidth}
        \centering
        \includegraphics[width=1.7in]{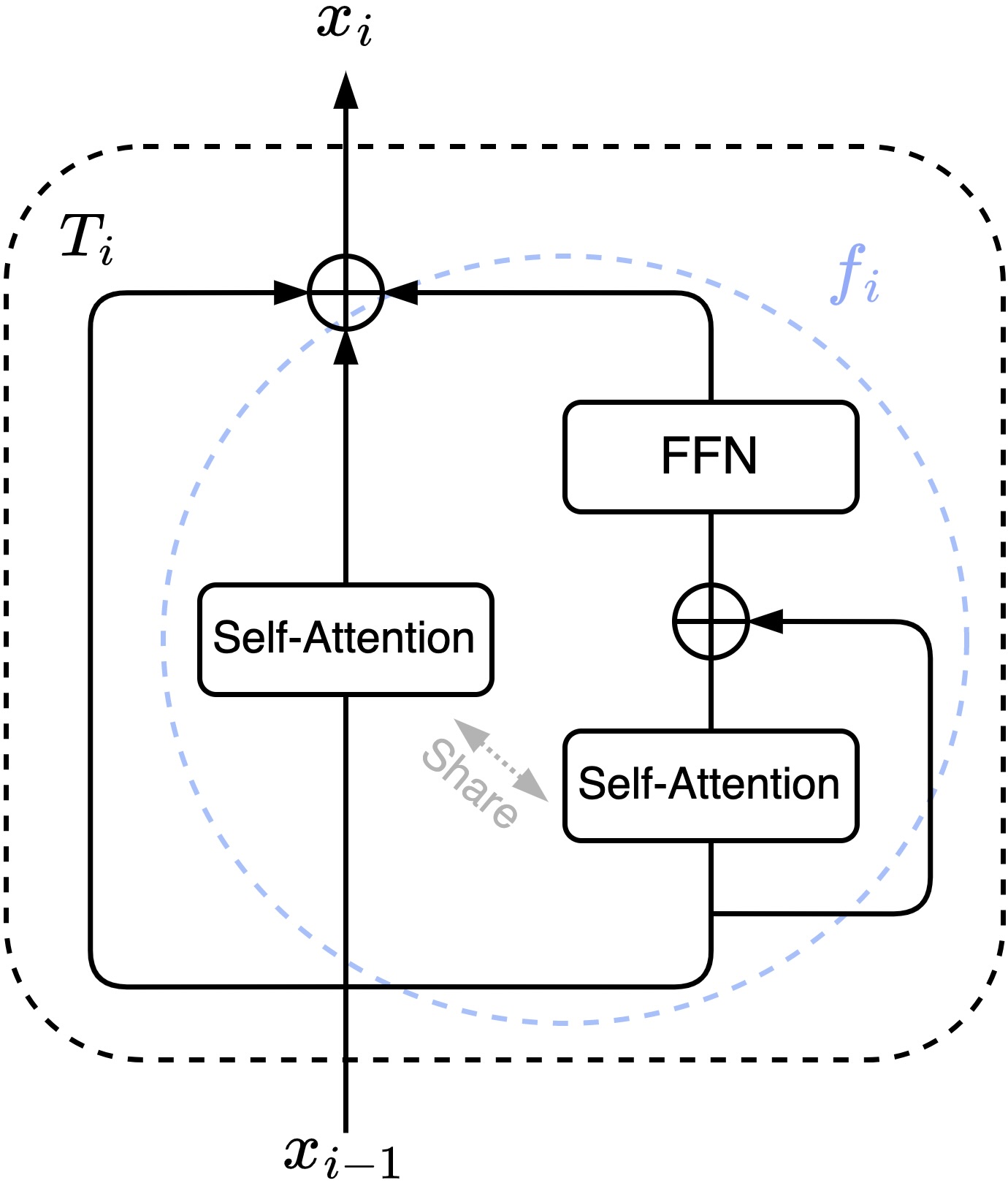}
        \caption{}
        \label{figue1b}
    \end{subfigure}
\caption{(a) Standard transformer form in modern ViTs is generally seen as a cascade of self-attention and FFN. (b) A three-path parallel form of transformer obtained by the equivalent transformation of (a).\vspace{-5mm}}
\label{transform}
\end{figure}

Residual connections~\cite{he2016deep} are universally adopted to bypass their sub-layers in ViTs, allowing data to flow from the previous layer directly to the subsequent layer. They are defined as the form
\begin{equation}
\label{residual}
\setlength{\abovedisplayskip}{3pt}
\setlength{\belowdisplayskip}{3pt}
x_{i}=g_i(x_{i-1}) +R_i(x_{i-1}),
\end{equation}
where the layer function $g_i$ and $R_i$ are typically identity and main building block.
In ViTs, we observe that nearly all the non-linear structures accord with the form of $R_i$ in Eq.~\ref{residual}, such as MHSA and FFN, and only linear structures exist between  $R_i$ and $R_{i+1}$ in most cases, so that the final feature fed into classifier can be seen as a linear combination of multi-path output. This key insightful observation inspires us that the ViTs can be viewed as a collection of many paths instead of a traditional single deep network. Specifically, we equivalently transform the traditional cascade of MHSA and FFN into three parallel paths in each transformer layer, as shown in Figure~\ref{transform}. Then, we utilize the identity connection in our new transformer form and further transform the ViT into an explicit multi-path ensemble network. Our ensemble network is equivalent to the traditional structure, which can be verified by mathematics and experiments, while the output of each path can be operated independently. 

%Many convolutional neural networks (CNNs), \eg, ResNet~\cite{he2016deep}, also employ residual connections. However, that non-linear activation functions exist between their residual blocks makes the models only implicit ensembles of numerous paths~\cite{veit2016residual} and difficult to be applied.

 In our ensemble view, each path performs two functions: the first is to provide the feature for the classifier directly, and the second is to provide the feature representation for subsequent longer paths. We propose new path combination and self-distillation to boost two functions separately to improve the performance of ViTs. We investigate the contribution of each path for the final prediction by analyzing the cosine distance and ablating different paths, and reveal that not all the paths are beneficial for the final results. Based on this observation, we design two simple and FLOPs-free path combination methods to optimize their combinations: path pruning which prunes underperforming paths, and EnsembleScale which re-weights the paths and makes the short paths focus on extracting high-quality representations for subsequent paths. Moreover, we discover that the model tends to enlarge the scales of the features of long paths to dilute the component of short paths, which increases the difficulty of optimization and raises the risk of divergence in deeper ViTs. Our EnsembleScale can make the model adjust the scale of EnsembleScale instead of features to alleviate this issue. According to the recent study of ViTs in frequency domain~\cite{wang2022anti,park2022how}, the low-pass filter property of self-attention weakens the expression of high-frequency signals. Our path combination methods can act as high-pass filters to remove partial useless low-frequency signals, and it achieves the goal of improving the first function of paths.

% Our path combination methods re-fuse the paths, but they cannot actively improve the representation learning of the paths.
% As shown in Figure~\ref{figure2}, the paths are optimized by the classification loss directly. However, not all the paths have enough capability in fitting the supervision from labels. The weak classification ability of some paths after training indicates that they are difficult to acquire benefits from classification loss. 
%Our path pruning and EnsembleScale optimize the combination of the paths.
To further improve the second function, that is, improving their representation utilized by the subsequent paths, we propose to  transfer knowledge between different paths by knowledge distillation (KD). 
% we introduce self-distillation to transfer the knowledge from the longer paths to the shorter paths. 
Thanks to the ensemble-like structure, we can perform self-distillation in a general teacher-student knowledge distillation way. We apply two types of distillation, prediction-logit distillation and hidden-state distillation, to allow the shorter paths to mimic the logit and feature relation of longer paths. Compared with traditional self-distillation methods~\cite{zhang2019your,kim2021self}, our method does not increase training parameters and memory cost.

The contributions of this paper are summarized as below.
\begin{itemize}
	\item We propose a novel view of ViTs, which illustrates that ViTs can be seen as a collection of paths, instead of a traditional single-path network. We can improve ViTs by optimizing the paths.
	\item Based on the proposed view, we investigate the contribution of different paths for the final prediction and find out that not all the paths are positive. We present path pruning and EnsembleScale to boost the ensemble performance. 
	\item To further enhance the representation ability of the paths, we design a self-distillation for ViTs. The teacher network and student network are appropriately selected from the paths, making the knowledge transfer among the paths effectively.
\end{itemize}
%-------------------------------------------------------------------------
\section{Related work}

\paragraph{Vision transformers.}
Vision transformer (ViT)~\cite{dosovitskiy2020vit} first introduces a pure Transformer backbone for image classification. ViT splits images into a sequence of tokens, and then adopts standard Transformer layers, consisting of Multi-Head Self-Attention (MHSA) and Feed-Forward Network (FFN), to model these tokens. Transformer, the core of ViT, and its sub-layers, MHSA and FFN, are improved to suit vision tasks by subsequent research and various remarkable ViT variants are proposed~\cite{touvron2021training,Xu_2021_ICCV,wang2021pyramid,chu2021conditional,wu2021cvt,jiang2021all,yu2022metaformer,liu2021swin,chu2021twins,yang2021focal,Dong_2022_CVPR,huang2021shuffle,li2022uniformer,chang2023making,yin2022avit,gao2023sparseformer}. For instance, PVT~\cite{wang2021pyramid} incorporates a spatial-reduction attention layer to achieve a high-resolution multi-scale design, favoring dense prediction tasks under limited computational cost. CVT~\cite{wu2021cvt} proposes a convolutional projection in the attention layer to combine the merits of CNNs and Transformers. Swin Transformer~\cite{liu2021swin} presents non-overlapping window partitions and restricts self-attention computation within windows to obtain linear computational complexity. Focal Transformer~\cite{yang2021focal} adopts focal self-attention to capture fine-grained local and coarse-grained global interactions. MetaFormer~\cite{yu2022metaformer} shows that replacing the self-attention with a spatial pooling operator can achieve competitive performance on many vision tasks and conclude the success of ViTs from the MetaFormer architecture. Those variants obey the basic ViT architecture, \ie, a stack of transformers containing residual sub-layers, which is the base of our ensemble view. Nearly no non-linear structures between adjacent residual blocks ensure that the final feature can be equivalently decoupled into multiple paths. 

Beyond image classification, ViT variants further inspire the application of transformer to other vision tasks, such as object detection~\cite{carion2020end,zhu2020deformable,zheng2020end,dai2021up}, semantic segmentation~\cite{wang2021max,wang2021end,zheng2021rethinking}, and self-supervised learning~\cite{chen2021mocov3,caron2021emerging,li2021efficient}.
\vspace{-3mm}
\paragraph{Ensemble.}
Neural network ensemble is a learning paradigm to collect a finite number of neural networks for the same task, originating from~\cite{hansen1990neural}. A neural network ensemble is normally constructed in two steps, training a number of component networks and combining the component predictions. The most classical methods of training component neural networks include Bagging~\cite{breiman1996bagging} and Boosting~\cite{schapire1990strength}. For combining the predictions of component neural networks, the most prevailing approaches are plurality voting or majority voting~\cite{hansen1990neural}, simple averaging~\cite{opitz1996actively}, and weighted averaging~\cite{perrone1992networks}. Zhou \etal~\cite{zhou2002ensembling} discuss the relationship between the ensemble and its component neural networks and uncovers that ensembling many of the components may be better than ensembling all of them. In this paper, the traditional ViTs are seen as an ensemble view and show ensemble-like behavior. We explore the contribution of each path for the vision task and improve the performance by deleting the weak components or introducing EnsembleScale to re-weight the components. Veit \etal ~\cite{veit2016residual} show that convolutional residual networks can be interpreted as a collection of many paths. The paths in~\cite{veit2016residual} are unrolled recursively from the bottom of the models, while our paths come from the linear combination of the top feature of the models. Moreover, the ensemble in~\cite{veit2016residual} is not an equivalent transformation since they neglect the non-linear structures between adjacent residual blocks in CNNs, which limits their practical application. 
\vspace{-3mm}
\paragraph{Knowledge distillation.}
Knowledge distillation~\cite{hinton2015distilling} transfers knowledge from a teacher model to a student model in a teacher-student framework. It has been widely studied in convolutional networks~\cite{DBLP:journals/corr/RomeroBKCGB14,passalis2018learning,komodakis2017paying,chen2020learning,gou2021knowledge,beyer2022knowledge}. Recently, several works develop distillation techniques for ViTs. DeiT~\cite{touvron2021training} applies a distillation token to transfer the knowledge from CNNs to transformers. MiniViT~\cite{zhang2022minivit} and TinyViT~\cite{wu2022tinyvit} adopt knowledge distillation to achieve lightweight ViTs. Manifold~\cite{jia2021efficient} proposes to excavate patch-level information to enhance ViT distillation. 

Besides general knowledge distillation, several works try to use the student network itself as a teacher, named self-knowledge distillation. BYOT~\cite{zhang2019your} boosts the low-level features by additional supervision from labels and the deepest layer. Xu \etal ~\cite{xu2019data} adopt different data distortions to deal with the same data and reduce their feature distance from a single network. CS-KD~\cite{yun2020regularizing} minimizes the KL divergence between predictive distributions from the same class. PS-KD~\cite{kim2021self} enhances the $k$-th epoch training model by incorporating the knowledge from the ${k-1}$-th epoch model. Although self-distillation methods avoid teacher networks, they increase other overhead, \eg, extra large-scale parameters~\cite{zhang2019your}, memory~\cite{kim2021self}, and computational cost~\cite{xu2019data,yun2020regularizing}.

%-------------------------------------------------------------------------

\vspace{-1mm}
\section{Method}
\vspace{-1mm}
The ensemble view of ViTs is first introduced in Sec.~\ref{3.1}, followed by the proposed path combination in Sec.~\ref{3.2} and self-distillation in Sec.~\ref{3.3}.
\subsection{The ensemble view of ViTs}
\label{3.1}
To make it easier to explain, we adopt the vanilla ViT~\cite{dosovitskiy2020vit} to instantiate a particular ensemble network. 
Consider a ViT network with $N$ transformer blocks $[T_i]_{i=1}^{N}$. For transformer $T_i$ consisting of two sub-layers, MHSA and FFN, $x_{i-1}$ and $x_i$ are defined as its input and output separately. Transformer is conventionally illustrated as in Figure~\ref{figure1a}, which is a cascade of MHSA and FFN with residual connections and a natural representation is written as
\begin{equation}
\label{eq_mhsa}
\setlength{\abovedisplayskip}{3pt}
\setlength{\belowdisplayskip}{3pt}
    x_{i-1}' = x_{i-1}+MHSA_i(x_{i-1}),
\end{equation}
\begin{equation}
\label{eq_ffn}
    x_i = x_{i-1}'+FFN_i(x_{i-1}').
\end{equation}
In a ViT, Eq.~\ref{eq_mhsa} and Eq.~\ref{eq_ffn} are alternately executed $N$ times and the output of the last transformer $x_N$ is fed into a classifier to generate the final prediction. We observe that there are two identity skip connections bridging the input and output of MHSA and FFN, respectively, which means that the input and output of MHSA can reach the tail of this transformer directly. Therefore, we can rearrange the transformer from a two-layer cascade form to a three-path parallel form as shown in Figure~\ref{figue1b}. The three paths include an identity skip connection, an MHSA layer, and an FFN followed by an MHSA layer. Two MHSA layers are weight sharing. The mathematical expression of Figure~\ref{figue1b} is
\begin{align}
%\begin{gathered}
\label{eq_new_transformer}
\setlength{\abovedisplayskip}{3pt}
\setlength{\belowdisplayskip}{3pt}
    x_{i} = &x_{i-1}+MHSA_i(x_{i-1}) \notag\\
          &+FFN_i(x_{i-1}+MHSA_i(x_{i-1})).
%\end{gathered}
\end{align}
The three paths in Figure~\ref{figue1b} correspond to the three terms in Eq.~\ref{eq_new_transformer}. In fact, Eq.~\ref{eq_new_transformer} is equivalent to the combination of Eq.~\ref{eq_mhsa} and Eq.~\ref{eq_ffn} by eliminating $x_{i-1}'$. We combine two parameterized paths into one network $f_i$ for convenience and thus the $i$-th transformer can be represented by
\begin{equation}
\setlength{\abovedisplayskip}{3pt}
\setlength{\belowdisplayskip}{3pt}
    x_{i} = x_{i-1}+f_i(x_{i-1}).
\end{equation}

\begin{figure}
\setlength{\abovecaptionskip}{0.cm}
\includegraphics[width=1.\linewidth]{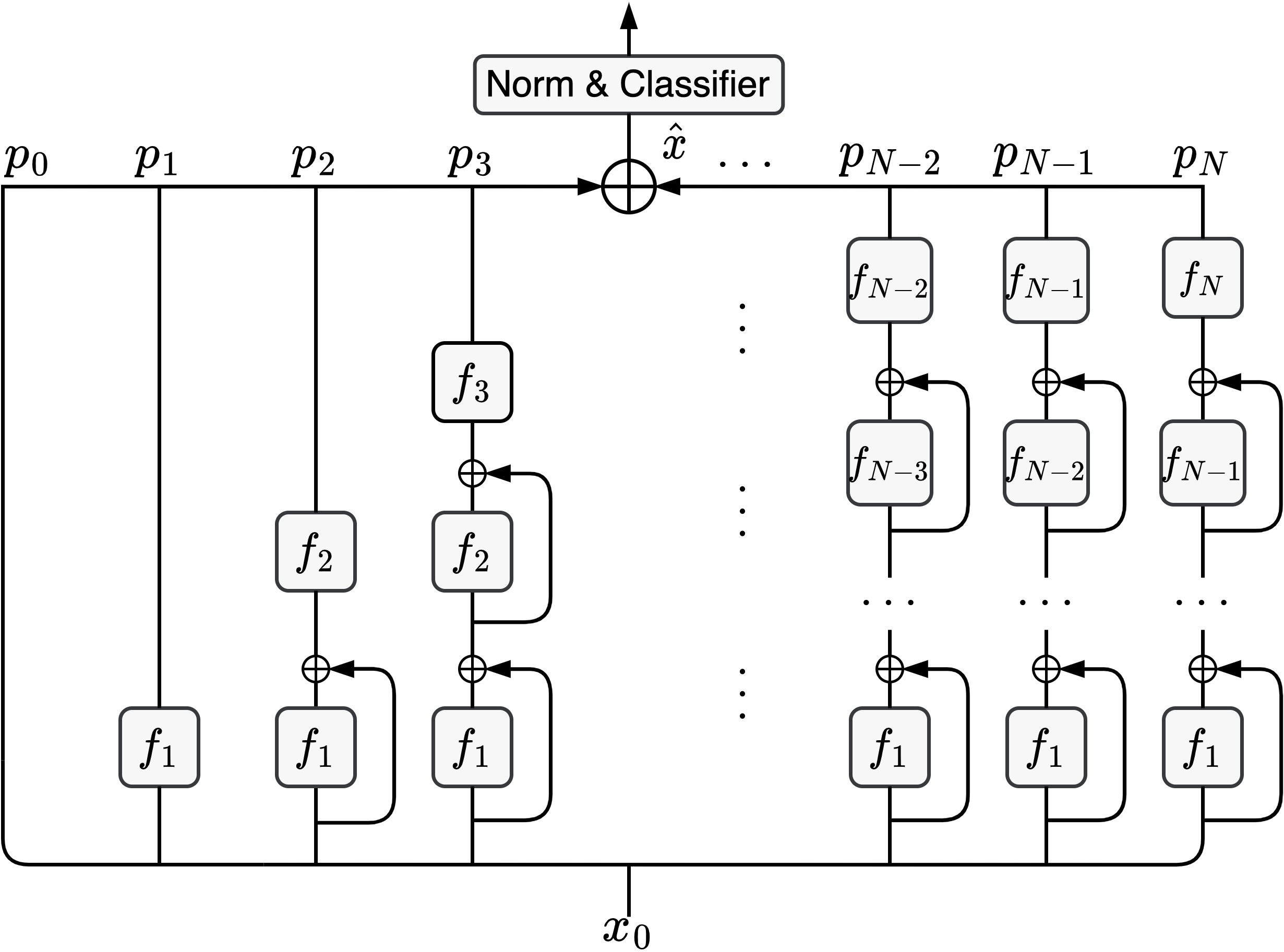}
\centering
\caption{Our ensemble view of ViTs.\vspace{-6mm}}
\label{figure2}
\end{figure}

Consider the output of the last transformer $x_{N}=x_{N-1}+f_N(x_{N-1})$, it can be seen as the a linear combination of $x_{N-1}$ and $f_N(x_{N-1})$. The term $x_{N-1}$ can be further decoupled into $x_{N-2}$ and $f_{N-1}(x_{N-2})$. In this recursive paradigm, we can unroll the $x_N$ into a linear number of terms, expanding one layer at each substitution step:
\begin{align}
\label{eq_ensemble}
\setlength{\abovedisplayskip}{3pt}
\setlength{\belowdisplayskip}{3pt}
    x_{N}&=x_{N-1}+f_N(x_{N-1})\notag\\
         &=x_{N-2}+f_{N-1}(x_{N-2})+f_N(x_{N-1})\\
         &\cdots\notag\\
         &=x_0+f_{1}(x_{0})+f_{2}(x_{1})+\cdots+f_N(x_{N-1}).\notag
\end{align}
As shown in Eq.~\ref{eq_ensemble}, we transform the top feature $x_N$ into a linear combination of $N+1$ terms. 
% Note that $x_n$ also can be decoupled into $2n+1$ terms if we do not combine two parameterized paths in Figure~\ref{figue1b}. 
The top feature $x_N$ is used to extract the class token (or average token), and then the class token (or average token) is input into a linear classifier to obtain the final predicted result. We regard the $N+1$ terms as $N+1$ paths and the final result as the ensemble of $N+1$ paths. Finally, a traditional ViT architecture is transformed to an ensemble view by the aforementioned transformation and analysis, and the ensemble view is illustrated in Figure~\ref{figure2}.

We denote the path corresponding to the term $f_i(x_{i-1})$ as $p_i$ ($i\in[0, N]$, where $p_0$ represents the term $x_0$). The network of $p_i$ is composed of the $i-1$ whole transformers $[T_j]_{j=1}^{i-1}$ and the parameterized sub-layers of transformer $T_i$, \ie, $p_i=f_i(T_{i-1}(\cdots T_1(x_0)))$. Obviously, the path with greater subscript contains more layers. The original $x_i$ can be denoted as the summation of the first $i$ paths in our ensemble form:
\begin{equation}
\setlength{\abovedisplayskip}{3pt}
\setlength{\belowdisplayskip}{3pt}
    x_{i} = \sum_{j=0}^i{p_j}.
\end{equation}
We define the ensemble feature fed into the classifier as $\hat{x}$, \ie, $\hat{x}=\sum_{i=0}^N{p_i}$. For a standard ViT, $\hat{x} = x_N$.

\begin{figure*}[t]
\setlength{\abovecaptionskip}{0.1cm}
    \begin{subfigure}{.33\textwidth}
        \centering
        \includegraphics[height=1.6in]{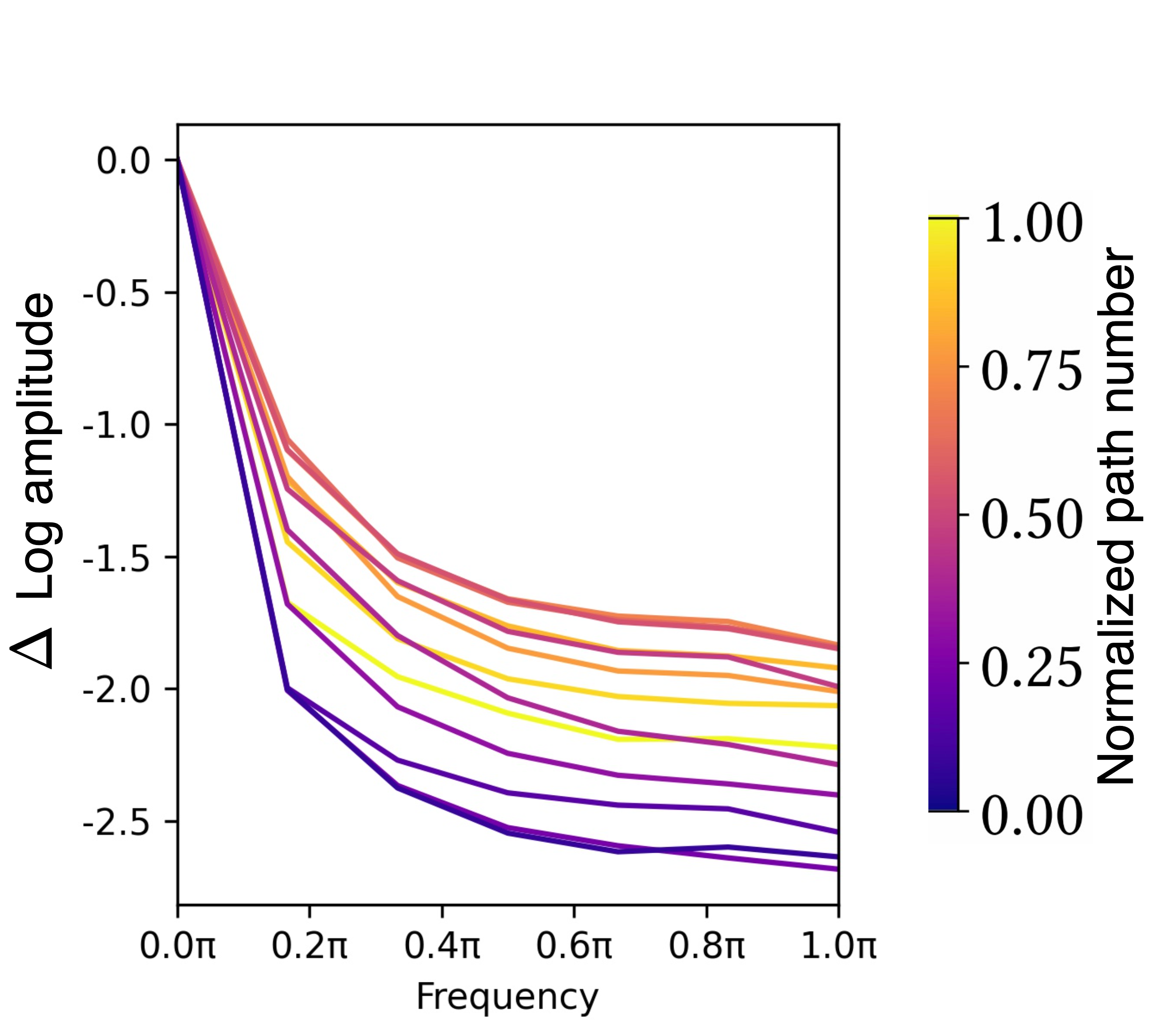}
        \caption{}
        \label{figure3c}
    \end{subfigure}
    \begin{subfigure}{.33\textwidth}
        \centering
        \includegraphics[height=1.4in]{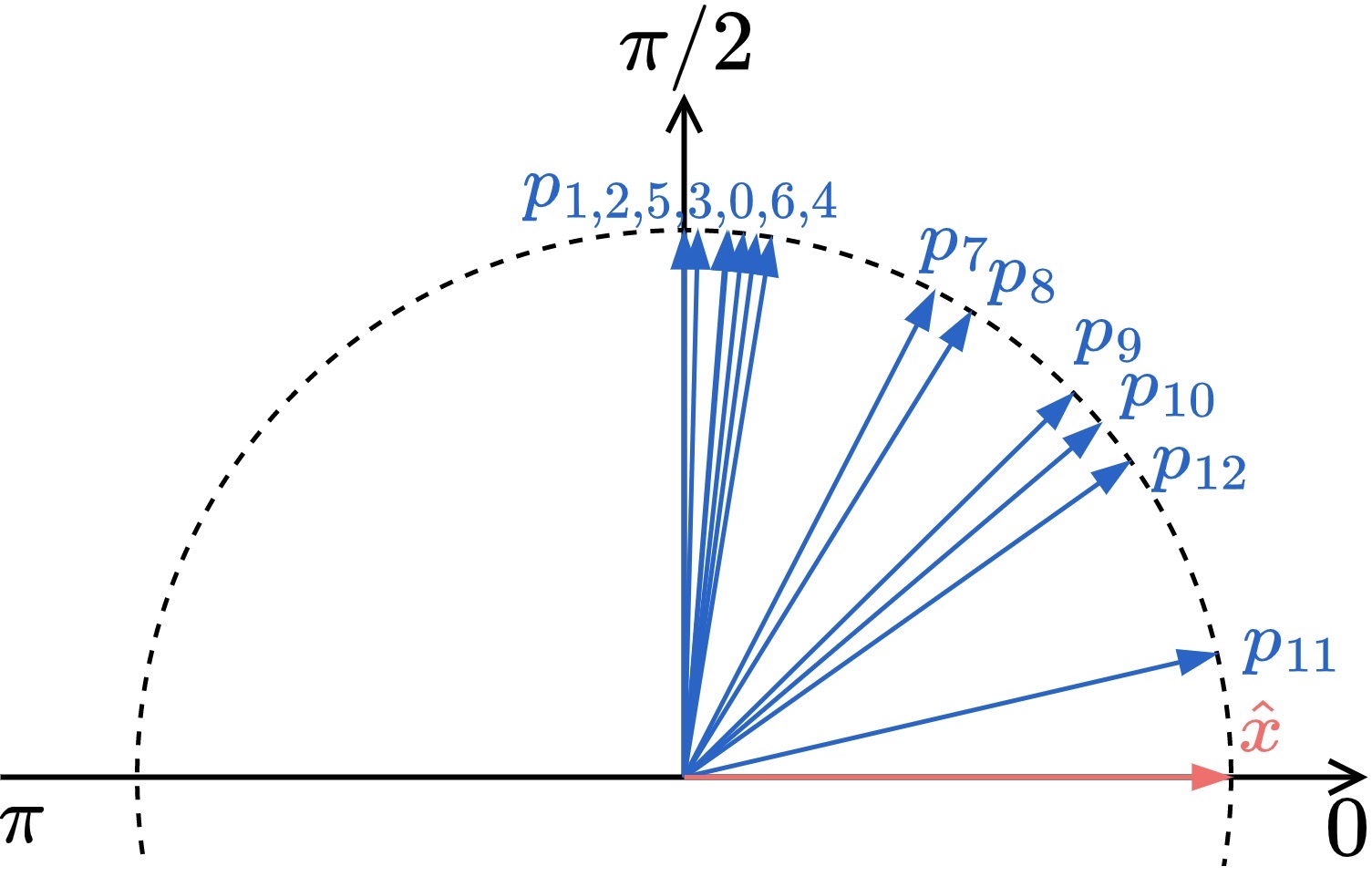}
        \caption{}
        \label{figure3a}
    \end{subfigure}
    \begin{subfigure}{.33\textwidth}
        \centering
        \includegraphics[height=1.4in]{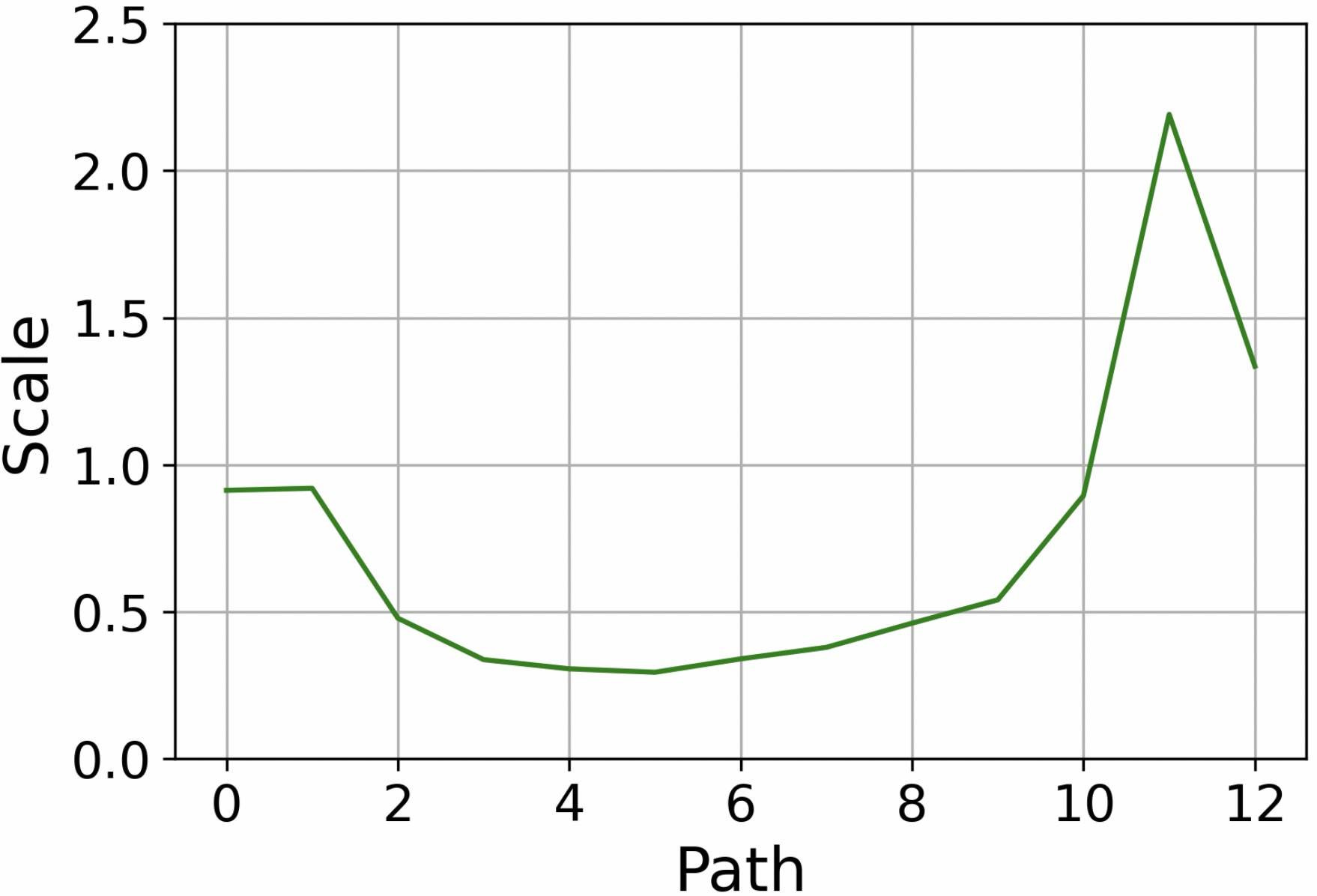}
        \caption{}
        \label{figure3b}
    \end{subfigure}
\caption{(a) Relative log amplitudes of Fourier transformed path features show the trend of the frequency of paths from raising to decline. This visualization refers to~\cite{park2022how}. (b) The cosine similarity projected in the interval $[0, \pi]$ between each path and ensemble feature $\hat{x}$. (c) The scales of the paths.\vspace{-3mm}}
\end{figure*}

Note that the new nodes $p_i$ in our view are not explicitly present in the traditional ViT architecture. \textbf{Our ensemble view provides a new perspective for ViTs, which involves developing ViT by optimizing the $\boldsymbol{N+1}$ paths.} We can process the paths to influence the final results. Moreover, the paths with different lengths also can be regarded as different frequency components of the final feature $\hat{x}$. Recent works~\cite{wang2022anti,park2022how,rao2021global,bai2022improving} have revealed the importance of frequency characteristics for ViTs. We show the Fourier analysis of paths in Figure~\ref{figure3c}. The frequency of paths show a trend from raising to declining. Many low-frequency components are concentrated in the short paths.  Adjusting the feature frequency also can be achieved by processing the paths.

% Above discussion is based on the most basic and simplest ViT paradigm~\cite{dosovitskiy2020vit}. However, most state-of-the-art ViT models adopt a hierarchical structure to utilize multi-scale features~\cite{wu2021cvt,wang2021pyramid,liu2021swin,chu2021twins,yang2021focal,Dong_2022_CVPR}. They split the transformers into multiple stages and insert a downsampling layer to reduce the resolution after each stage. The downsampling layer is composed of linear layers generally including a normalization layer and a linear layer (or convolutional layer) without non-linear layers, so the hierarchical ViTs can also be transformed into our ensemble form. 
% The normalization layer needs to be particularly expounded. ViTs widely employ LayerNorm~\cite{ba2016layer} as the normalization layer in the downsampling layer. In our ensemble view, each path calculates the standard-deviation independently, which makes the forward propagation of our ensemble view and the standard view not equivalent unless synchronizing the standard-deviation. We conduct extensive experiments to study this issue. It is discovered that the model can adapt the asynchronous standard-deviation and the performance of asynchronous standard-deviation slightly outperforms the baseline when we train the models in the ensemble form from scratch. Therefore, we do not cope with the asynchronous standard-deviation problem in the subsequent study.

The above discussion pertains to the most basic and simple ViT paradigm~\cite{dosovitskiy2020vit}. However, most state-of-the-art ViT models adopt a hierarchical structure to utilize multi-scale features~\cite{wu2021cvt,wang2021pyramid,liu2021swin,chu2021twins,yang2021focal,Dong_2022_CVPR}. These models split transformers into multiple stages and insert downsampling layers to reduce resolution after each stage. The downsampling layer typically comprises linear layers which include a normalization layer and a linear or convolutional layer, without non-linear layers. As a result, hierarchical ViTs can also be transformed into our ensemble form. 
The normalization layer needs to be particularly expounded. ViTs widely employ LayerNorm~\cite{ba2016layer} as the normalization layer in the downsampling layer. In our ensemble view, each path calculates the standard deviation independently, which makes the forward propagation of our ensemble view and the standard view not equivalent, unless synchronizing the standard deviations. We conduct extensive experiments to study this issue and find that the model can adapt to asynchronous standard deviation. Additionally, the performance of asynchronous standard deviation slightly outperforms synchronous standard deviation when we train the models in the ensemble form from scratch. Therefore, we directly adopt the asynchronous standard deviation in subsequent studies. More details about our ensemble form of hierarchical ViTs are in our Appendix~\ref{a 8}.

For non-hierarchical ViTs~\cite{dosovitskiy2020vit,touvron2021training,jiang2021all}, all the intermediate variables of our ensemble form are also computed in the standard form. Hence, the FLOPs and throughput of our ensemble form are equal to the standard form. However, for hierarchical ViTs, the downsampling layers need to be implemented in each path in our ensemble view, which incurs significant computational cost. To address this issue, we apply the strategy of ``summation before downsampling". Unlike computing the output of all the paths before combining, we sum all the existing paths once encountering downsampling layers. In this way, the computational complexity from additional downsampling layers is linear to the number of downsampling layers rather than linear to the number of transformers. In the next subsection, we can see that the extra computational complexity can be reduced further. The practical FLOPs increase is negligible.

Figure \ref{figure2} makes clear the processes of the well-known ViT in a novel ensemble view, where the data flow along multiple paths with different depth to form the ensemble prediction and each path also acts as the bottom network of the longer paths. 
Consequently, each path performs two functions: the first is to provide the feature for the classifier, and the second is to extract semantic representation for subsequent long paths. Based on these observations, we formulate the following questions: is ensembling all the paths the optimal solution? If not, how to optimize the combination of paths? Besides the classification supervision, do we have a better manner to improve the representation of the paths?

\subsection{Path combination}
\label{3.2}
% In Figure \ref{figure2}, the ViT is seen as a collection of many paths with different length and each path is also the bottom network of the longer paths. Therefore, each path performs two functions: the first is to provide the feature for the classifier, and the second is to extract semantic representation for subsequent long paths. 
% For short paths, such as $P_0$ and $P_1$, their networks are composed of a few layers leading to poor semantic representations. However, their output is also collected by the final prediction and supervised by the label directly. 

The ensemble feature $\hat{x}=\sum_{i=0}^N{p_i}$ is fed into the linear classifier to produce prediction (the extraction of class token or average pooling is ignored for brevity). 
This procedure can be simplified as $y = \hat{x}w+b$, where $y$ is the predicted result and $w$ and $b$ are the weight and the bias of the classifier. $w$ can be regarded as a set of $c$ vectors, \ie, $[w_1, w_2, \cdots, w_c]$, where $c$ is the number of classes. Assuming that the class $gt$ corresponding to $w_{gt}$ is the ground truth, the model is expected to make $\hat{x}w_{gt} > \hat{x}w_{i}$ ($i\in [1,c]$ and $i\ne gt$). From the view of high dimensional space, it is equivalent to minimize the angle between $\hat{x}$ and $w_{gt}$. We use $\hat{x}$ to approximately substitute $w_{gt}$ to measure the classification ability of each path. A toy experiment is conducted by taking a ViT-S model with 12 transformer layers pre-trained on the ImageNet-1K training set. We normalize the output of the paths and project the cosine similarity between each path and $\hat{x}$ in the interval $[0, \pi]$ as shown in Figure~\ref{figure3a}. The features from short paths are nearly orthogonal to $\hat{x}$, reflecting their weak classification ability. 
% We argue whether incorporating those low-level feature from short paths benefits the final prediction. 

Short paths with weak classification ability can play two roles in the combination of final prediction: (i) providing low-level information to revise the results of long paths; (ii) acting as the noise to be diminished by enlarging the scale of the features of long paths or sparsifying classifier weight $w$.
%We normalize the path output and computing the cosine similarity between each path and $\hat{x}$ to measure the component of $\hat{x}$.
% Thanks to the final prediction being equivalent to the linear combination of paths, we can ablate the paths to investigate their influence on the final prediction. 
We conduct another experiment to evaluate this ViT-S with different combinations of the paths on the validation set and show the results in Table~\ref{table path}.
Note that the model is not fine-tunned even though it has been changed due to path ablation. Table \ref{table path} presents that several long paths contribute the majority of accuracy. In particular, the last three paths, \ie, $p_{10}$, $p_{11}$, and $p_{12}$ attain 99.9\% of the baseline accuracy. Moreover, we notice that the combination of $[p_i]_{i=2}^{12}$ is slightly superior to the baseline even though the parameters of normalization layer and classifier are optimized for the baseline. According to these evidences, we recognize that the features of short paths do not necessarily benefit the final prediction. 

We use the $l_1$ norm to calculate the scales of $[p_i]_{i=0}^{12}$ and plot them in Figure \ref{figure3b}. The scale curve presents an obvious escalating trend and reaches the peak at $p_{11}$, which indicates that the model fights against the noisy features by enhancing the scale of the main components. However, this way increases the difficulty of optimization and makes the model unstable with more layers. We argue that this is one of the factors causing performance saturation in deeper ViTs~\cite{touvron2021going,wang2022anti}.

\begin{table}[t]
\small
\begin{center}
\begin{tabular}{c|c}
            \toprule
            The combination of paths & Top-1 Acc (\%) \\
            \midrule
            Baseline ($p_{12}$, $p_{11}$, $\cdots$, $p_{0}$) & 80.31* \\
            \midrule
            $p_{12}$ & 78.70 \\
            $p_{11}$ & 79.70 \\
            $p_{12}$, $p_{11}$ & 80.14 \\
            $p_{12}$, $p_{11}$, $p_{10}$ & 80.22 \\
            $p_{12}$, $p_{11}$, $\cdots$, $p_{2}$ & 80.36 \\
            \bottomrule
        \end{tabular}
\end{center}
\vspace{-3mm}
\caption{Inference the pre-trained ViT-S model containing 12 transformer layers with different combinations of paths. The baseline model contains the full paths. *: Using average pooling to replace class token like~\cite{touvron2021going}.\vspace{-5mm}}
\label{table path}
\end{table}

\begin{figure}
\setlength{\abovecaptionskip}{0.1cm}
\includegraphics[width=1.\linewidth]{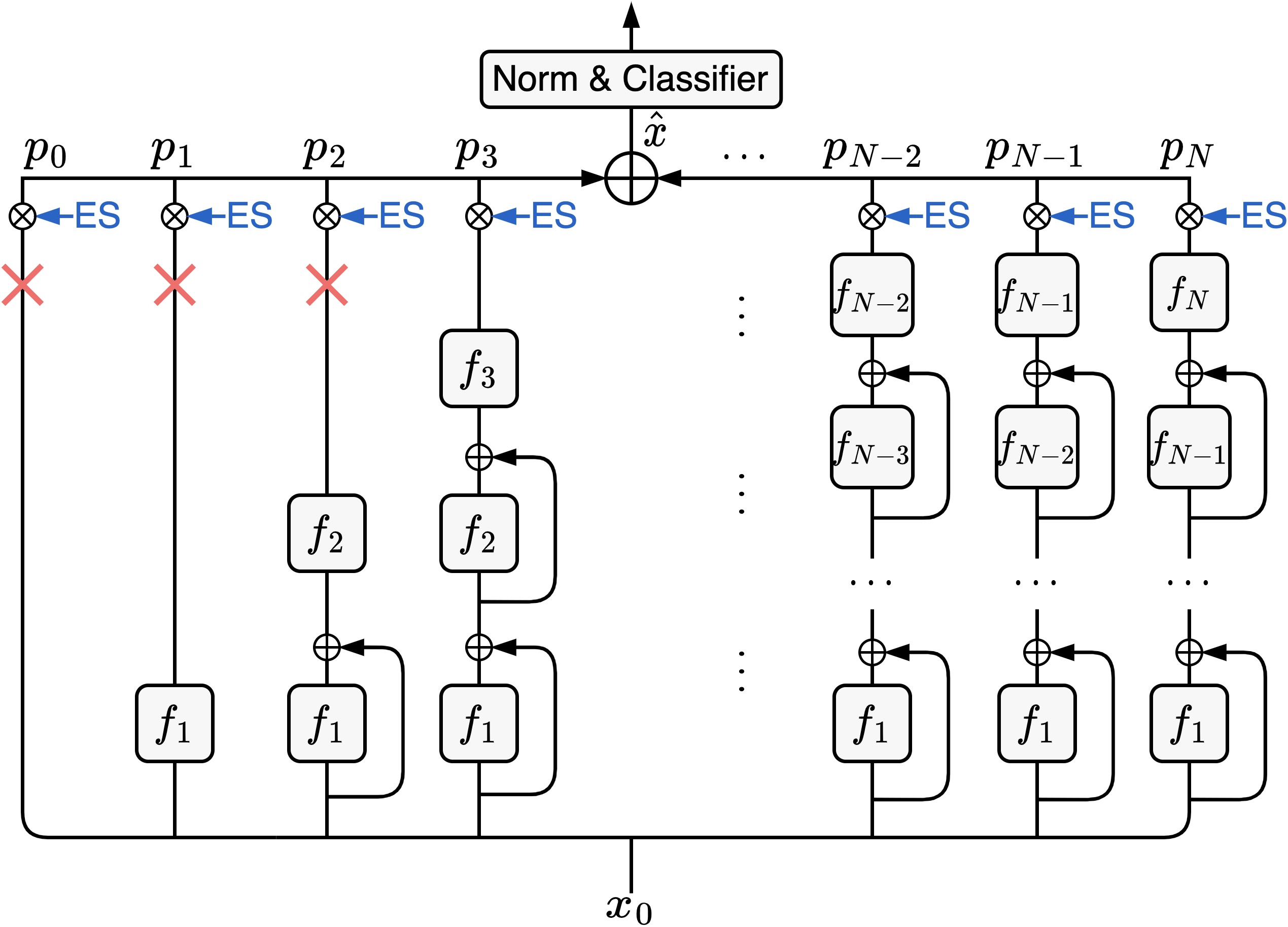}
\centering
\caption{We propose two schemes, path pruning and EnsembleScale, to optimize path combination. \textcolor[RGB]{255,102,102}{\textbf{$\times$}} represents cutting out the corresponding path and \textcolor[RGB]{0,102,156}{ES} is short for EnsembleScale.\vspace{-5mm}}
\label{figure4}
\end{figure}

Built on the above analysis, we propose path pruning and EnsembleScale to optimize the path combination. 
\paragraph{Path pruning.} We prune some short paths as shown in Figure \ref{figure4} and force the shallow transformers to focus on extracting low-level semantic representation for subsequent layers. % 不影响优化，不等同于在原始结构图上改
We restate that the paths in our ensemble view are not explicit in the traditional view of Figure~\ref{figure1a}. Path pruning is not equivalent to removing residual connections in Figure \ref{figure1a}. We only prevent the ensemble prediction from combining the predictions from short paths. The shallow layers can be optimized well by residual connection in the remaining paths.

% For non-hierarchical ViTs, path pruning does not impact the computational complexity. For hierarchical ViTs, path pruning can reduce the extra FLOPs from downsampling layers. 
Path pruning has different effects on the computational complexity of non-hierarchical and hierarchical ViTs. Non-hierarchical ViTs remain unaffected by path pruning, whereas for hierarchical ViTs, it can help reduce the additional FLOPs required by downsampling layers.
Take Swin-T~\cite{liu2021swin} as an example of the latter. Swin-T needs to compute three additional downsampling layers in the ensemble form. If we remove the first three paths, the downsampling layer following stage 1 is not needed. The FLOPs of the ensemble form are equal to those of the standard form when only the last two paths are saved.

\paragraph{EnsembleScale.} We propose EnsembleScale which is a per-channel multiplication of the vector produced by each path to adaptively re-weight the combination of the paths. Formally, EnsembleScale is a multiplication by a diagonal matrix on the output of each path, denoted as 
\begin{equation}
\setlength{\abovedisplayskip}{3pt}
\setlength{\belowdisplayskip}{3pt}
    \hat{x} = \sum_{i=0}^{N}{diag(\lambda_{i,1}, \cdots,\lambda_{i,d})\times p_i},
\end{equation}
where the parameters $\lambda_{i,j}$ are learnable weights and $d$ is the number of channels. We initialize the EnsembleScale progressively from $10^{-5}$ to $1.0$ based on the analysis of short paths. EnsembleScale can be regarded as a soft path pruning. Another functionality of EnsembleScale is that the model can adjust the scale of paths by EnsembleScale rather than features in case that the scale of the feature expands with depth, which can help ViTs go deeper.

% We can also explain the effectiveness of our path combination strategies through frequency domain. The ensemble network decouples the final feature into the output features of multiple paths with different length. The frequency of these output features is from high to low with the increasing length. Therefore, the output features of the paths can be regarded as the different frequency components of the final feature of a ViT model. Image classification task prefers to low-frequency signals. Our path combination methods filter out high-frequency components and consequently work for image classification.

The effectiveness of our path combination strategies comes from two folds. First, the features generated by short paths focus on providing low-level representation for subsequent longer paths instead of minimizing the classification error. Second, 
% We explain the effectiveness of our path combination strategies from the perspective of frequency domain. The frequency of these output features is from high to low with increasing length. 
from the perspective of frequency domain, our path combination methods mainly filter out useless low-frequency signals, amounting to amplifying the effect of high-frequency signals. Recent works~\cite{wang2022anti,park2022how,rao2021global,bai2022improving} validate that the high-frequency components are generally overwhelmed in ViTs and appropriately compensating for them can boost the performance.

\subsection{Self-distillation}
\label{3.3}
% The paths are trained jointly and have the same optimized object with different depths. 
Path combination solves the combination of the paths, but it cannot actively optimize the training of the paths. To enhance their representation, we introduce self-distillation to transfer knowledge from the longer paths to the shorter paths.  
% According to two functions of the paths mentioned in Sec.~\ref{3.2}, \ie, providing features for classifier and extracting representation for subsequent paths, 
Two types of distillation are considered, \ie, prediction-logit distillation and hidden-state distillation.

\paragraph{Prediction-logit distillation.}
Given two paths, the path with the deeper network is selected as the teacher $p_t$ and the other as the student $p_s$. Then,  the classifier of the overall model is employed to extract their logits, to avoid introducing additional networks. Finally, we force  $p_s$ to imitate $p_t$ to regularize the student path. This is achieved by a Kullback-Leibler divergence loss as below:
\begin{equation}
\setlength{\abovedisplayskip}{3pt}
\setlength{\belowdisplayskip}{3pt}
    \mathcal{L}_{pl} = KL(\sigma(\frac{Cls(p_s)}{T}) || \sigma(\frac{Cls(p_t)}{T})),
\end{equation}
where $\sigma(\cdot)$ is the Softmax function, $T$ is a temperature value controlling the smoothness of the logits, and $Cls$ denotes the classifier including a LayerNorm and a linear layer. We do not update the parameters of the classifier in this loss.

\paragraph{Hidden-state distillation.}
We compute the relations among tokens in $p_s$ and $p_t$, respectively, and obtain two relation matrices defined by $R_i = \textit{softmax}(p_i\cdot p_i^T/\sqrt{d})$. The hidden-state distillation loss based on relation matrices is achieved by another Kullback-Leibler divergence loss:
\begin{equation}
\setlength{\abovedisplayskip}{3pt}
\setlength{\belowdisplayskip}{3pt}
    \mathcal{L}_{hidden} = KL(R_s || R_t).
\end{equation}

In experiments, we find that a large representation gap between the teacher and the student leads to inferior performance. Thus, a distance constant $\Delta$ is set to constrain teacher-student pair, \ie, $p_{i+\Delta}$ teaching $p_i$. $\Delta$ is set to $2$ by default. The final distillation objective function is formulated as
\begin{equation}
\setlength{\abovedisplayskip}{3pt}
\setlength{\belowdisplayskip}{3pt}
\label{overall loss}
    \resizebox{0.85\hsize}{!}{
    $\mathcal{L}_{kd} = \sum_{i=s}^{N-\Delta}{\alpha_i\mathcal{L}_{pl}(p_i, p_{i+\Delta})+\beta_i\mathcal{L}_{hidden}(p_i, p_{i+\Delta})},$
    }
\end{equation}
where $s$ represents the subscript of the starting path, and $\alpha_i$ and $\beta_i$ are hyperparameters to balance the loss.

\section{Experiments}
\label{4}
In this section, we report our experimental results related to path combination and self-distillation.
\paragraph{Experimental settings.}
Our method is verified on two representative ViT models, DeiT~\cite{touvron2021training} and Swin~\cite{liu2021swin}. All the models are trained on the ImageNet~\cite{deng2009imagenet} with 1.28M training images and 50K validation images from 1,000 classes. The image resolution in training and inference is $224\times 224$. All the models are trained for a total of 300 epochs, while the batch size is set to 1,024. The augmentation and regularization strategies follow the original papers of DeiT and Swin.

\subsection{Path combination}
\paragraph{Main results.} 
Our results of path combination on DeiT and Swin are summarized in Table~\ref{table deit}. We report top-1 accuracy, the number of parameters, and FLOPs under different path combination settings. The number of parameters and FLOPs of this paper are measured by Fvcore\footnote{\url{https://github.com/facebookresearch/fvcore}}. For DeiT, as our analysis, the components of short paths serve more as useless low-frequency information, thus pruning them can boost the performance. For instance, the performance of DeiT-S is improved by 0.4\% when only keeping $p_7$-$p_{12}$. The gain comes for free without any additional parameters or FLOPs. The optimal path combination $p_7$-$p_{12}$ is consistent with the cosine similarity analysis in Figure~\ref{figure3a} which shows that $p_0$-$p_{6}$ are nearly orthogonal to the final ensemble feature. Our EnsembleScale re-weights the path combination and achieves better performance than path pruning in most cases. 
% The frequency analysis in Figure~\ref{figure6a} and Figure~\ref{figure6b} illustrates that our path pruning and EnsembleScale increase the proportion of high-frequency signals.

Our methods also work well with Swin Transformers. Both path pruning and EnsembleScale can bring improvement, which demonstrates that our method is effective for diverse ViT models. The FLOPs increase compared with the baseline is from the transformation of ensemble form. Our EnsembleScale does not augment the FLOPs and path pruning can diminish this FLOPs increase due to less utilization of downsampling layers.
% The FLOPs increase slightly due to the additional downsampling layers.

\begin{figure}
\setlength{\abovecaptionskip}{0.1cm}
\begin{subfigure}{.235\textwidth}
        \centering
        \includegraphics[width=1.5in]{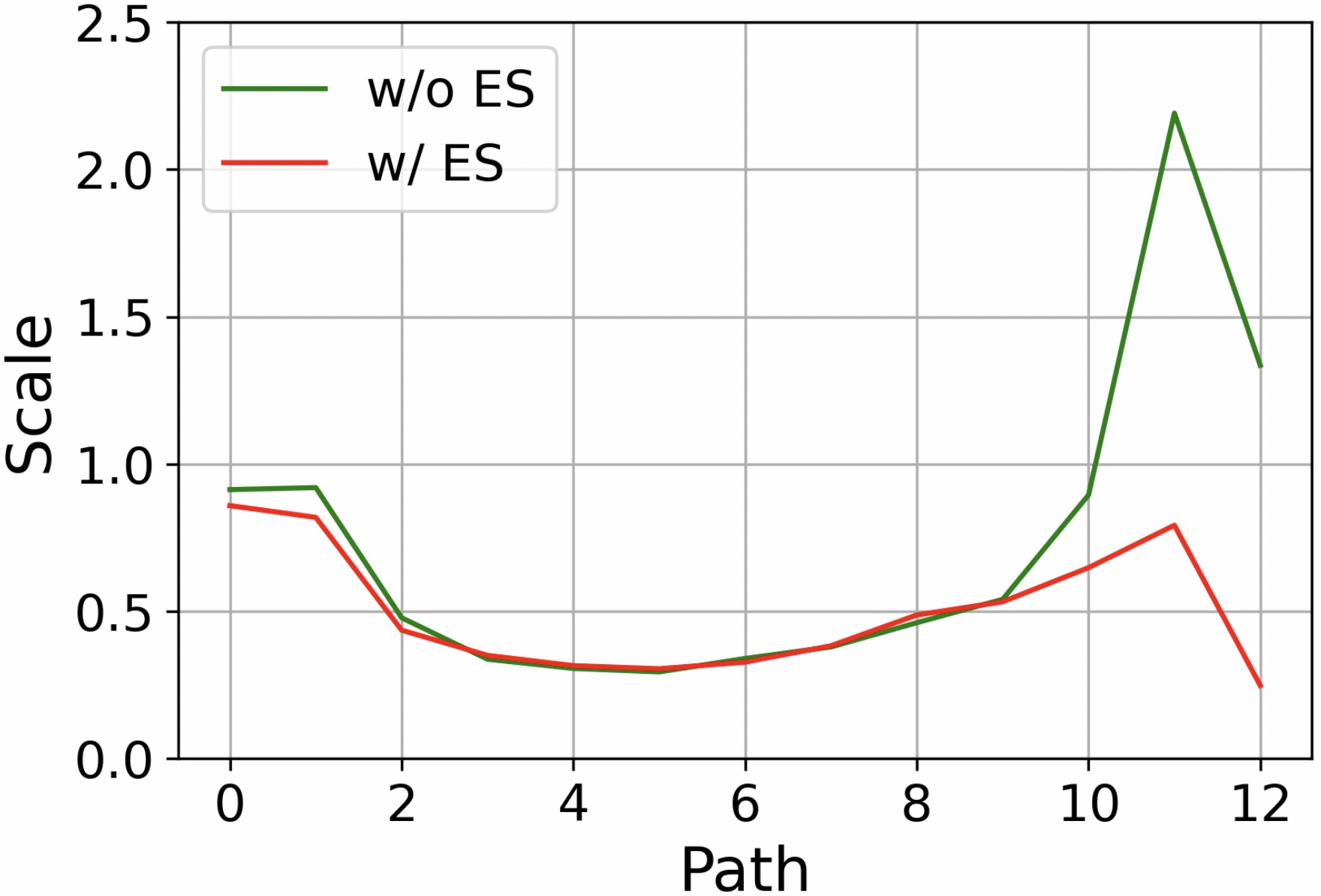}
        \caption{}
        \label{figure5a}
    \end{subfigure}
    \begin{subfigure}{.235\textwidth}
        \centering
        \includegraphics[width=1.5in]{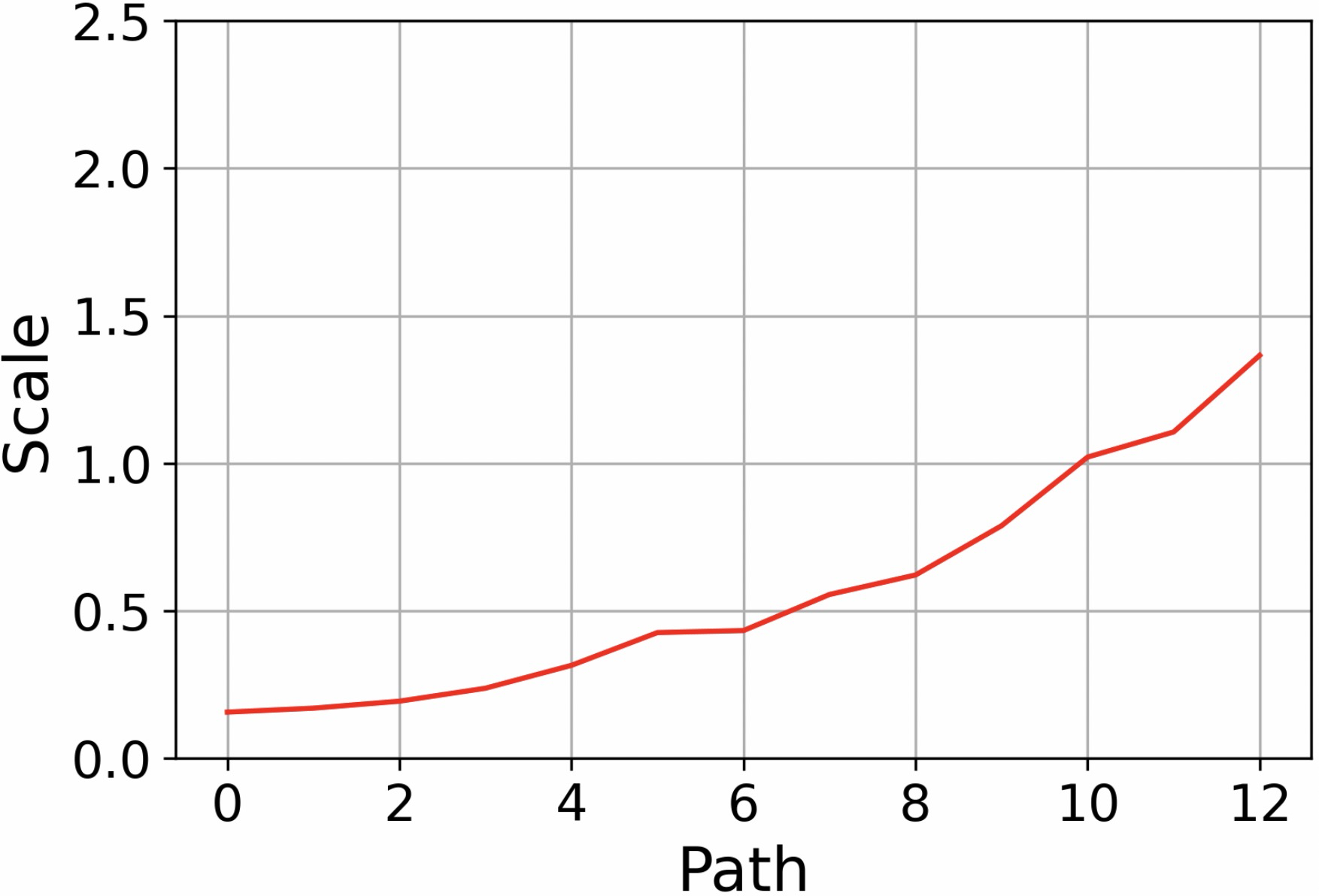}
        \caption{}
        \label{figure5b}
    \end{subfigure}
\vspace{-5mm}
\caption{(a) The scales of the paths with and without EnsembleScale. (b) The scales of EnsembleScale.}
\vspace{-1mm}
\label{figure5}
\end{figure}

% \paragraph{Improving convergence at depth.}
\paragraph{Making ViTs go deeper.}
We visualize the feature scales of the paths on DeiT-S and DeiT-S with EnsembleScale separately in Figure~\ref{figure5a}. It can be seen that DeiT-S has to expand the scales of long paths to suppress weak features from short paths, while the model with EnsembleScale can adjust EnsembleScale (Figure~\ref{figure5b}) to balance the weight of the paths so that the feature scales of long paths do not need to be extremely large.

We argue that large scale is one of the reasons for collapsed deep ViTs and our EnsembleScale can mitigate this issue. We experiment with more transformer layers on DeiT-S with EnsembleScale to evaluate the stability in Table~\ref{table stable}. Note that all the hyper-parameters of ``DeiT-S+ES'' are the same as the vanilla DeiT-S~\cite{touvron2021training}. From this table, EnsembleScale is able to converge with more layers without saturating too early. Moreover, EnsembleScale brings more improvement when more transformer layers are introduced. For example, EnsembleScale enhances 18-layer DeiT-S by 1.1\%, which is far more than the 0.5\% improvement it brings on 12-layer DeiT-S. Finally, these experiments support our hypothesis that it is the behavior of ViTs enlarging the scale of long paths to dilute the components of short paths that impedes ViTs going deeper. Prior works~\cite{touvron2021going,wang2022anti,zhou2021deepvit,wang2022anti} also explore the deeper ViTs but explain and solve this issue from other perspectives. We think that our work unravels a new factor of degraded deep ViTs, and EnsembleScale can actually be complementary to previous works.
% Our work unravels a new interpretation of degraded deep ViTs and our EnsembleScale is complementary to them.

%\paragraph{Visualization and frequency domain analysis.}
%We visualize the feature maps of DeiT-S and DeiT-S with EnsembleScale in Figure~\ref{figure6}. Their patterns are significantly different. DeiT-S adopts high values/low values to represent foreground/background. In contrast, DeiT-S with EnsembleScale applies low values/high values to represent foreground/background. Another important point is that the DeiT-S contains remarkably more high-frequency signals than DeiT with EnsembleScale. The paths of our ensemble form also can be described as frequency decouple. The frequency is from high to low with the increasing length of paths. In image classification task, our path combination strategies amount to low-pass filters as shown in Figure~\ref{figure6}. If we reverse the weights of the paths, \ie, assigning more weights to short paths, our path combination can become high-pass filters to boost some fine-grained downstream tasks. We will further explore this application in the future.

\paragraph{Efficient dynamic ViTs.}
Our ensemble view can be leveraged to design efficient dynamic ViTs. We observe that many images can be predicted accurately using only a few paths, while a small fraction of difficult images require processing through the entire network. We apply a simple approach to achieve a dynamic ViT in this experiment. We use two groups of EnsembleScale, denoted as $ES_1$ and $ES_2$, to combine the first seven paths and all the paths, respectively, and generate two predicted features, $\hat{x_1}$ and $\hat{x_2}$. We apply $\hat{x_1}$ to produce the initial prediction and terminate the inference process once a sufficiently confident prediction is generated. The score output by the classifier serves as the measure of confidence. Our results are shown in Table \ref{dynamic}.
After adding $ES_1$ and $ES_2$ and finetuning, the accuracy of DeiT-S increases to 80.0\% when evaluating the validation set using all the paths. Subsequently, we apply our dynamic ViT to this network, resulting in a 25\% reduction in FLOPs and an accuracy of 79.8\% which is the same as the accuracy of the original DeiT-S.
% demonstrate that we can reduce FLOPs by 25\% while only suffering from a 0.2\% decrease in accuracy. 
These findings indicate that our ensemble view has vast potential to achieve efficient ViT design.

The number of data processed by different stages of our dynamic ViT is shown in Table~\ref{table count}. It is observed that approximately 51.6\% ``easy" image are processed using the first 7 paths with an accuracy of 93.1\%, while the remaining 48.4\% ``hard" data require  the whole network. 
Our dynamic ViT is the simplest implementation. We believe that more effective and efficient dynamic ViT can be achieved based on our ensemble form.

\paragraph{Transfer learning.}
It is important to evaluate our method on other datasets with transfer learning in order to measure the generalization ability of our method. The transfer learning tasks are performed by finetuning the model on CIFAR-100~\cite{krizhevsky2009learning} and CIFAR-10~\cite{krizhevsky2009learning} as shown in Table~\ref{table transfer}. For finetuning, we use the same training setting as ImageNet-1K pre-training. Our EnsembleScale achieves superior results on both CIFAR-100 and CIFAR-10, demonstrating excellent generalization ability.

\begin{table}[t]
\small
\begin{center}
\resizebox{!}{4.2cm}{
\begin{tabular}{c|c|c|c|c|c}
            \toprule
            Model & Path & ES & \# Params & FLOPs & Top-1 Acc (\%)  \\
            \midrule
            % \multirow{5}{*}{DeiT-T} & - & & 5.72M & 1.26G & 72.2 (Baseline)\\
            % & $P_6$ - $P_{12}$ & & 5.72M & 1.26G & 72.4 \\
            % & $P_7$ - $P_{12}$ & & 5.72M & 1.26G & 72.5 \\
            % & $P_8$ - $P_{12}$ & & 5.72M & 1.26G & 72.3 \\
            % & $P_0$ - $P_{12}$ & \checkmark & 5.72M & 1.26G & 72.5 \\
            % \midrule
            \multirow{5}{*}{DeiT-S} & - & & 22.05M & 4.58G & 79.8 (Baseline)\\
            & $p_6$ - $p_{12}$ & & 22.05M & 4.58G & 80.1 \\
            & $p_7$ - $p_{12}$ & & 22.05M & 4.58G & 80.2 \\
            & $p_8$ - $p_{12}$ & & 22.05M & 4.58G & 80.1 \\
            & $p_0$ - $p_{12}$ & \checkmark & 22.06M & 4.58G & 80.3\\
            \midrule
            \multirow{5}{*}{DeiT-B} & - & & 86.57M & 17.58G & 81.8 (Baseline)\\
            & $p_6$ - $p_{12}$ & & 86.57M & 17.58G & 82.0 \\
            & $p_7$ - $p_{12}$ & & 86.57M & 17.58G & 82.2 \\
            & $p_8$ - $p_{12}$ & & 86.57M & 17.58G & 82.2\\
            & $p_0$ - $p_{12}$ & \checkmark & 86.58M & 17.58G & 82.3\\
            \bottomrule
            \multirow{5}{*}{Swin-T} & - & & 28.29M & 4.51G & 81.3 (Baseline)\\
            & $p_6$ - $p_{12}$ & & 28.29M & 4.56G & 81.5 \\
            & $p_8$ - $p_{12}$ & & 28.29M & 4.56G & 81.5 \\
            % & $P_{11}$ - $P_{12}$ & & 28.29M & 4.51G & 81.2 \\
            & $p_0$ - $p_{12}$ & \checkmark & 28.29M & 4.68G & 81.5\\
            \midrule
            \multirow{5}{*}{Swin-S} & - & & 49.61M & 8.77G & 83.0 (Baseline)\\
            & $p_6$ - $p_{24}$ & & 49.61M & 8.83G & 83.2 \\
            & $p_8$ - $p_{24}$ & & 49.61M & 8.83G &  83.2 \\
            % & $P_8$ - $P_{24}$ & & 49.61M & 8.83G &  83.2\\
            & $p_0$ - $p_{24}$ & \checkmark & 40.61M & 8.95G & 83.3\\
            \midrule
            \multirow{5}{*}{Swin-B} & - & & 87.77M & 15.47G & 83.5 (Baseline)\\
            & $p_6$ - $p_{24}$ & & 87.77M & 15.57G & 83.7 \\
            & $p_7$ - $p_{24}$ & & 87.77M & 15.57G & 83.7 \\
            % & $P_8$ - $P_{24}$ & & 87.77M & 15.57G & 83.7\\
            & $p_0$ - $p_{24}$ & \checkmark & 87.78M & 15.78G & 83.8\\
            \bottomrule
\end{tabular}}
\vspace{-3mm}
\end{center}
\caption{Applying path pruning and EnsembleScale to DeiT-(Small, Base) and Swin-(Tiny, Small, Base). The top-1 accuracy, the number of parameters and FLOPs are reported under different settings. ES is short for EnsembleScale.}
\label{table deit}
\end{table}

% \begin{figure}
% \begin{subfigure}{.235\textwidth}
%         \centering
%         \includegraphics[width=1.5in]{figure6a.jpeg}
%         \caption{The baseline models w/ and w/o path pruning.}
%         \label{figure6a}
%     \end{subfigure}
%     \begin{subfigure}{.235\textwidth}
%         \centering
%         \includegraphics[width=1.5in]{figure6b.jpeg}
%         \caption{The baseline models w/ and w/o EnsembleScale.}
%         \label{figure6b}
%     \end{subfigure}
% \vspace{-5mm}
% \caption{Relative log amplitude of Fourier transformed final features.\vspace{-5mm}}
% \label{figure6}
% \end{figure}

\begin{table}[t]
\small
\begin{center}
\begin{tabular}{c|c|c}
            \toprule
            Depth & DeiT-S & DeiT-S + ES   \\
            \midrule
            18 & 80.1 & 81.2 \\
            24 & 78.9$\dagger$ & 81.6 \\
            \bottomrule
        \end{tabular}
\end{center}
\vspace{-4mm}
\caption{Evaluting convergence at depth on DeiT-S. ES is short for EnsembleScale. The accuracy of DeiT is reported by~\cite{touvron2021going}. $\dagger$: failed before the end of training.}
\label{table stable}
\end{table}

\begin{table}[t]
\small
\begin{center}
\begin{tabular}{c|c|c}
            \toprule
            Model & FLOPs(G) & Accuracy(\%)    \\
            \midrule
            DeiT-S & 4.58 & 79.8  \\
            DeiT-S (finetuning) & 4.58 & 80.0  \\
            Dynamic DeiT-S & 3.42(-25\%) & 79.8  \\
            \bottomrule
\end{tabular}
\end{center}
\vspace{-4mm}
\caption{We compare the our dynamic DeiT-S with the original DeiT-S model, reporting their respective top-1 accuracy and FLOPs.}
\label{dynamic}
\vspace{-1mm}
\end{table}

\begin{table}[t]
\small
\begin{center}
\begin{tabular}{c|c|c}
            \toprule
             & First 7 paths & All the paths    \\
            \midrule
            No. of images & 25824 (51.65\%) & 24176 (48.35\%)  \\
            Accuracy(\%) & 93.1 & 65.5  \\
            \bottomrule
        \end{tabular}
\end{center}
\vspace{-3mm}
\caption{The number of data processed by different stages of our dynamic ViT.}
\label{table count}
\end{table}

\begin{table}[t]
\small
\begin{center}
\begin{tabular}{c|c|c|c}
            \toprule
            Model & ImageNet-1K & CIFAR-100 & CIFAR-10   \\
            \midrule
            DeiT-S & 79.8 & 85.9 & 98.1 \\
            DeiT-S with ES & 80.3 & 87.0 & 98.6\\
            \bottomrule
        \end{tabular}
\end{center}
\vspace{-4mm}
\caption{Results in transfer learning. We compare DeiT-S with EnsembleScale to DeiT-S on CIFAR-100 and CIFAR-10.\vspace{-3mm}}
\label{table transfer}
\end{table}

\subsection{Self-distillation}

\paragraph{Main results.}
We train DeiT and Swin Transformer with our self-distillation on ImageNet-1K and evaluate them on the validation set of ImageNet-1K. As shown in Table~\ref{table sd}, both prediction-logit distillation and hidden-state distillation on DeiT and Swin can improve performance compared to the baselines, which verifies the effectiveness of our distillation methods. We also combine  self-distillation with path combination and the performance can be further enhanced. Our path combination and self-distillation are built on our ensemble view and achieve promising results for nearly FLOPs-free and parameter-free, presenting the significant potential of the ensemble view. 

\vspace{-2mm}
\paragraph{Teacher selection.}
Given a path, we investigate how to select its teacher path. We set a constant $\Delta$ to control the distance between the teacher path and the student path. Different $\Delta$ are evaluated on DeiT-S in Table~\ref{table short}. Different from the experience in CNNs~\cite{zhang2019your}, employing the deepest features as teachers cannot bring improvement. Our method achieves the best performance when $\Delta$ is 2. 

% We visualize the cosine similarity among all the paths in Figure~\ref{figure7}. Adjacent paths normally have similar semantic information, so adjacent paths are easy to inject knowledge into the student path.
\vspace{-1mm}
\paragraph{Do we need to distill short paths?}
In path combination, we halt the data flow from short paths to the final prediction. We conduct experiments to verify the effectiveness of distilling short paths. Table~\ref{table teacher} illustrates the performance of different starting paths in Eq~\ref{overall loss}. For instance, ``$p_2$'' represents the paths $p_2$-$p_{12}$ involve in distillation. We can see that the optimal result is from $p_5$ and distilling extremely short paths cannot improve performance. 

% \paragraph{Effects of different loss.}

\begin{table}[t]
\small
\begin{center}
\resizebox{!}{1.57cm}{
\begin{tabular}{c|c|c|c|c}
            \toprule
            \multirow{2}{*}{Model} & \multicolumn{4}{c}{Method}   \\
            & Base & PL & PL+HS & PL+HS+Path combination \\
            \midrule
            DeiT-S & 79.8 & 80.2 & 80.6 & 81.0\\
            \midrule
            DeiT-B & 81.8 & 82.1 & 82.4 & 82.7 \\
            \midrule
            Swin-T & 81.3 & 81.7 & 82.1  & 82.3\\
            \midrule
            Swin-S & 83.0 & 83.4 &83.6  & 83.8\\
            \midrule
            Swin-B & 83.5 & 84.3 &84.0  & 84.2\\
            \bottomrule
        \end{tabular}
}
\end{center}
\vspace{-3mm}
\caption{Applying our self-distillation to DeiT and Swin on ImageNet-1K. PL and HS are short for prediction-logit distillation and hidden-state distillation, respectively. We also report the results of the combination of self-distillation and path combination in the last column.}
\vspace{-2mm}
\label{table sd}
\end{table}

\begin{table}[t]
\small
\begin{center}
\begin{tabular}{c|c|c|c|c}
            \toprule
            $\Delta$ & 1 & 2 & 3 & 4    \\
            \midrule
            Top-1 Acc (\%)& 80.5 & 81.0 & 80.6 & 80.4 \\
            \bottomrule
        \end{tabular}
\end{center}
\vspace{-3mm}
\caption{Performance evaluation on different values of $\Delta$ on DeiT-S.}
\vspace{-2mm}
\label{table short}
\end{table}

\begin{table}[t]
\small
\begin{center}
\begin{tabular}{c|c|c|c|c|c}
            \toprule
            Starting path & $p_0$ & $p_2$ & $p_4$ & $p_5$ & $p_6$    \\
            \midrule
            Top-1 Acc (\%)& 80.6 & 80.8 & 80.8 & 81.0 & 80.8 \\
            \bottomrule
        \end{tabular}
\end{center}
\vspace{-3mm}
\caption{Performance evaluation on different starting path on DeiT-S.}
\vspace{-6mm}
\label{table teacher}
\end{table}

\section{Conclusion}
% In this paper, we revisit ViTs and propose a new view showing that they can be seen as an ensemble of many paths with different length. The transformation from standard form to our ensemble form is equivalent, which allows us to manipulate the paths for various purposes. We investigate the functionalities of the paths and argue that the short paths do not benefit the final prediction. To optimize the path combination, we propose to re-weight the paths from an ensemble learning perspective. Our path combination can also help the ViTs go deeper and have the effect of modulating frequency. Moreover, we present a self-distillation method to transfer knowledge from long paths to short paths to enhance the representation of the paths. 

% In the future, we will further explore how to utilize the paths beyond the methods proposed in this paper, such as tuning the path components for downstream vision tasks. Furthermore, it is meaningful to investigate whether our ensemble view supports NLP networks. We hope that this work can inspire more future research to design and optimize the ViTs in an ensemble view.

In this paper, we revisit ViTs and propose a novel ensemble view that shows ViTs as an ensemble of multiple paths with varying lengths. We demonstrate that the transformation from the standard form to our ensemble form is equivalent, enabling us to manipulate the paths for different purposes. Through our investigation, we argue that short paths do not benefit the final prediction and propose new strategies to re-weight the paths from an ensemble learning perspective to optimize the path combination. Our method can also help ViTs go deeper and modulate frequency. Moreover, we introduce a self-distillation method to transfer knowledge from long paths to short paths to enhance the representation of the paths.

In the future, we plan to explore further ways to utilize the paths beyond the methods proposed in this paper, such as tuning the path components for downstream vision tasks. Furthermore, it would be worthwhile to investigate whether our ensemble view supports NLP networks. We hope that this work inspires more research in the future to design and optimize ViTs using an ensemble view.

\section*{Acknowledgement}
This project is supported by the National Research Foundation, Singapore under its NRFF Award NRF-NRFF13-2021-0008, and Mike Zheng Shou's Start-Up Grant from NUS. Shuning Chang was supported by Alibaba Group through Alibaba Research Intern Program.
% {\small
% \bibliographystyle{ieee_fullname}
% \bibliography{egbib}

\begin{thebibliography}{10}\itemsep=-1pt

\bibitem{ba2016layer}
Jimmy~Lei Ba, Jamie~Ryan Kiros, and Geoffrey~E Hinton.
\newblock Layer normalization.
\newblock {\em arXiv preprint arXiv:1607.06450}, 2016.

\bibitem{bai2022improving}
Jiawang Bai, Li Yuan, Shu-Tao Xia, Shuicheng Yan, Zhifeng Li, and Wei Liu.
\newblock Improving vision transformers by revisiting high-frequency
  components.
\newblock {\em arXiv preprint arXiv:2204.00993}, 2022.

\bibitem{beyer2022knowledge}
Lucas Beyer, Xiaohua Zhai, Am{\'e}lie Royer, Larisa Markeeva, Rohan Anil, and
  Alexander Kolesnikov.
\newblock Knowledge distillation: A good teacher is patient and consistent.
\newblock In {\em Proceedings of the IEEE/CVF Conference on Computer Vision and
  Pattern Recognition}, pages 10925--10934, 2022.

\bibitem{breiman1996bagging}
Leo Breiman.
\newblock Bagging predictors.
\newblock {\em Machine learning}, 24(2):123--140, 1996.

\bibitem{carion2020end}
Nicolas Carion, Francisco Massa, Gabriel Synnaeve, Nicolas Usunier, Alexander
  Kirillov, and Sergey Zagoruyko.
\newblock End-to-end object detection with transformers.
\newblock In {\em European conference on computer vision}, pages 213--229.
  Springer, 2020.

\bibitem{caron2021emerging}
Mathilde Caron, Hugo Touvron, Ishan Misra, Herv{\'e} J{\'e}gou, Julien Mairal,
  Piotr Bojanowski, and Armand Joulin.
\newblock Emerging properties in self-supervised vision transformers.
\newblock In {\em Proceedings of the IEEE/CVF International Conference on
  Computer Vision}, pages 9650--9660, 2021.

\bibitem{chang2023making}
Shuning Chang, Pichao Wang, Ming Lin, Fan Wang, David~Junhao Zhang, Rong Jin,
  and Mike~Zheng Shou.
\newblock Making vision transformers efficient from a token sparsification
  view.
\newblock In {\em Proceedings of the IEEE/CVF Conference on Computer Vision and
  Pattern Recognition}, pages 6195--6205, 2023.

\bibitem{chen2020learning}
Hanting Chen, Yunhe Wang, Chang Xu, Chao Xu, and Dacheng Tao.
\newblock Learning student networks via feature embedding.
\newblock {\em IEEE Transactions on Neural Networks and Learning Systems},
  32(1):25--35, 2020.

\bibitem{chen2021mocov3}
Xinlei Chen*, Saining Xie*, and Kaiming He.
\newblock An empirical study of training self-supervised vision transformers.
\newblock {\em arXiv preprint arXiv:2104.02057}, 2021.

\bibitem{chu2021twins}
Xiangxiang Chu, Zhi Tian, Yuqing Wang, Bo Zhang, Haibing Ren, Xiaolin Wei,
  Huaxia Xia, and Chunhua Shen.
\newblock Twins: Revisiting the design of spatial attention in vision
  transformers.
\newblock {\em Advances in Neural Information Processing Systems},
  34:9355--9366, 2021.

\bibitem{chu2021conditional}
Xiangxiang Chu, Zhi Tian, Bo Zhang, Xinlong Wang, Xiaolin Wei, Huaxia Xia, and
  Chunhua Shen.
\newblock Conditional positional encodings for vision transformers.
\newblock {\em arXiv preprint arXiv:2102.10882}, 2021.

\bibitem{dai2021up}
Zhigang Dai, Bolun Cai, Yugeng Lin, and Junying Chen.
\newblock Up-detr: Unsupervised pre-training for object detection with
  transformers.
\newblock In {\em Proceedings of the IEEE/CVF Conference on Computer Vision and
  Pattern Recognition}, pages 1601--1610, 2021.

\bibitem{deng2009imagenet}
Jia Deng, Wei Dong, Richard Socher, Li-Jia Li, Kai Li, and Li Fei-Fei.
\newblock Imagenet: A large-scale hierarchical image database.
\newblock In {\em 2009 IEEE conference on computer vision and pattern
  recognition}, pages 248--255. Ieee, 2009.

\bibitem{Dong_2022_CVPR}
Xiaoyi Dong, Jianmin Bao, Dongdong Chen, Weiming Zhang, Nenghai Yu, Lu Yuan,
  Dong Chen, and Baining Guo.
\newblock Cswin transformer: A general vision transformer backbone with
  cross-shaped windows.
\newblock In {\em Proceedings of the IEEE/CVF Conference on Computer Vision and
  Pattern Recognition (CVPR)}, pages 12124--12134, June 2022.

\bibitem{dosovitskiy2020vit}
Alexey Dosovitskiy, Lucas Beyer, Alexander Kolesnikov, Dirk Weissenborn,
  Xiaohua Zhai, Thomas Unterthiner, Mostafa Dehghani, Matthias Minderer, Georg
  Heigold, Sylvain Gelly, Jakob Uszkoreit, and Neil Houlsby.
\newblock An image is worth 16x16 words: Transformers for image recognition at
  scale.
\newblock {\em ICLR}, 2021.

\bibitem{gao2023sparseformer}
Ziteng Gao, Zhan Tong, Limin Wang, and Mike~Zheng Shou.
\newblock Sparseformer: Sparse visual recognition via limited latent tokens.
\newblock {\em arXiv preprint arXiv:2304.03768}, 2023.

\bibitem{gou2021knowledge}
Jianping Gou, Baosheng Yu, Stephen~J Maybank, and Dacheng Tao.
\newblock Knowledge distillation: A survey.
\newblock {\em International Journal of Computer Vision}, 129(6):1789--1819,
  2021.

\bibitem{hansen1990neural}
Lars~Kai Hansen and Peter Salamon.
\newblock Neural network ensembles.
\newblock {\em IEEE transactions on pattern analysis and machine intelligence},
  12(10):993--1001, 1990.

\bibitem{he2016deep}
Kaiming He, Xiangyu Zhang, Shaoqing Ren, and Jian Sun.
\newblock Deep residual learning for image recognition.
\newblock In {\em Proceedings of the IEEE conference on computer vision and
  pattern recognition}, pages 770--778, 2016.

\bibitem{hinton2015distilling}
Geoffrey Hinton, Oriol Vinyals, Jeff Dean, et~al.
\newblock Distilling the knowledge in a neural network.
\newblock {\em arXiv preprint arXiv:1503.02531}, 2(7), 2015.

\bibitem{huang2021shuffle}
Zilong Huang, Youcheng Ben, Guozhong Luo, Pei Cheng, Gang Yu, and Bin Fu.
\newblock Shuffle transformer: Rethinking spatial shuffle for vision
  transformer.
\newblock {\em arXiv preprint arXiv:2106.03650}, 2021.

\bibitem{jia2021efficient}
Ding Jia, Kai Han, Yunhe Wang, Yehui Tang, Jianyuan Guo, Chao Zhang, and
  Dacheng Tao.
\newblock Efficient vision transformers via fine-grained manifold distillation.
\newblock {\em arXiv preprint arXiv:2107.01378}, 2021.

\bibitem{jiang2021all}
Zi-Hang Jiang, Qibin Hou, Li Yuan, Daquan Zhou, Yujun Shi, Xiaojie Jin, Anran
  Wang, and Jiashi Feng.
\newblock All tokens matter: Token labeling for training better vision
  transformers.
\newblock {\em Advances in Neural Information Processing Systems}, 34, 2021.

\bibitem{kim2021self}
Kyungyul Kim, ByeongMoon Ji, Doyoung Yoon, and Sangheum Hwang.
\newblock Self-knowledge distillation with progressive refinement of targets.
\newblock In {\em Proceedings of the IEEE/CVF International Conference on
  Computer Vision}, pages 6567--6576, 2021.

\bibitem{komodakis2017paying}
Nikos Komodakis and Sergey Zagoruyko.
\newblock Paying more attention to attention: improving the performance of
  convolutional neural networks via attention transfer.
\newblock In {\em ICLR}, 2017.

\bibitem{krizhevsky2009learning}
Alex Krizhevsky, Geoffrey Hinton, et~al.
\newblock Learning multiple layers of features from tiny images.
\newblock 2009.

\bibitem{li2021efficient}
Chunyuan Li, Jianwei Yang, Pengchuan Zhang, Mei Gao, Bin Xiao, Xiyang Dai, Lu
  Yuan, and Jianfeng Gao.
\newblock Efficient self-supervised vision transformers for representation
  learning.
\newblock {\em arXiv preprint arXiv:2106.09785}, 2021.

\bibitem{li2022uniformer}
Kunchang Li, Yali Wang, Peng Gao, Guanglu Song, Yu Liu, Hongsheng Li, and Yu
  Qiao.
\newblock Uniformer: Unified transformer for efficient spatiotemporal
  representation learning.
\newblock {\em arXiv preprint arXiv:2201.04676}, 2022.

\bibitem{liu2021swin}
Ze Liu, Yutong Lin, Yue Cao, Han Hu, Yixuan Wei, Zheng Zhang, Stephen Lin, and
  Baining Guo.
\newblock Swin transformer: Hierarchical vision transformer using shifted
  windows.
\newblock In {\em Proceedings of the IEEE/CVF International Conference on
  Computer Vision}, pages 10012--10022, 2021.

\bibitem{opitz1996actively}
David~W Opitz and Jude~W Shavlik.
\newblock Actively searching for an effective neural network ensemble.
\newblock {\em Connection Science}, 8(3-4):337--354, 1996.

\bibitem{park2022how}
Namuk Park and Songkuk Kim.
\newblock How do vision transformers work?
\newblock In {\em International Conference on Learning Representations}, 2022.

\bibitem{passalis2018learning}
Nikolaos Passalis and Anastasios Tefas.
\newblock Learning deep representations with probabilistic knowledge transfer.
\newblock In {\em Proceedings of the European Conference on Computer Vision
  (ECCV)}, pages 268--284, 2018.

\bibitem{perrone1992networks}
Michael~P Perrone and Leon~N Cooper.
\newblock When networks disagree: Ensemble methods for hybrid neural networks.
\newblock Technical report, Brown Univ Providence Ri Inst for Brain and Neural
  Systems, 1992.

\bibitem{rao2021global}
Yongming Rao, Wenliang Zhao, Zheng Zhu, Jiwen Lu, and Jie Zhou.
\newblock Global filter networks for image classification.
\newblock {\em Advances in Neural Information Processing Systems}, 34:980--993,
  2021.

\bibitem{DBLP:journals/corr/RomeroBKCGB14}
Adriana Romero, Nicolas Ballas, Samira~Ebrahimi Kahou, Antoine Chassang, Carlo
  Gatta, and Yoshua Bengio.
\newblock Fitnets: Hints for thin deep nets.
\newblock In {\em 3rd International Conference on Learning Representations,
  {ICLR} 2015, San Diego, CA, USA, May 7-9, 2015, Conference Track
  Proceedings}, 2015.

\bibitem{schapire1990strength}
Robert~E Schapire.
\newblock The strength of weak learnability.
\newblock {\em Machine learning}, 5(2):197--227, 1990.

\bibitem{touvron2021training}
Hugo Touvron, Matthieu Cord, Matthijs Douze, Francisco Massa, Alexandre
  Sablayrolles, and Herv{\'e} J{\'e}gou.
\newblock Training data-efficient image transformers \& distillation through
  attention.
\newblock In {\em International Conference on Machine Learning}, pages
  10347--10357. PMLR, 2021.

\bibitem{touvron2021going}
Hugo Touvron, Matthieu Cord, Alexandre Sablayrolles, Gabriel Synnaeve, and
  Herv{\'e} J{\'e}gou.
\newblock Going deeper with image transformers.
\newblock In {\em Proceedings of the IEEE/CVF International Conference on
  Computer Vision}, pages 32--42, 2021.

\bibitem{veit2016residual}
Andreas Veit, Michael~J Wilber, and Serge Belongie.
\newblock Residual networks behave like ensembles of relatively shallow
  networks.
\newblock {\em Advances in neural information processing systems}, 29, 2016.

\bibitem{wang2021max}
Huiyu Wang, Yukun Zhu, Hartwig Adam, Alan Yuille, and Liang-Chieh Chen.
\newblock Max-deeplab: End-to-end panoptic segmentation with mask transformers.
\newblock In {\em Proceedings of the IEEE/CVF Conference on Computer Vision and
  Pattern Recognition}, pages 5463--5474, 2021.

\bibitem{wang2022anti}
Peihao Wang, Wenqing Zheng, Tianlong Chen, and Zhangyang Wang.
\newblock Anti-oversmoothing in deep vision transformers via the fourier domain
  analysis: From theory to practice.
\newblock {\em arXiv preprint arXiv:2203.05962}, 2022.

\bibitem{wang2021pyramid}
Wenhai Wang, Enze Xie, Xiang Li, Deng-Ping Fan, Kaitao Song, Ding Liang, Tong
  Lu, Ping Luo, and Ling Shao.
\newblock Pyramid vision transformer: A versatile backbone for dense prediction
  without convolutions.
\newblock In {\em Proceedings of the IEEE/CVF International Conference on
  Computer Vision}, pages 568--578, 2021.

\bibitem{wang2021end}
Yuqing Wang, Zhaoliang Xu, Xinlong Wang, Chunhua Shen, Baoshan Cheng, Hao Shen,
  and Huaxia Xia.
\newblock End-to-end video instance segmentation with transformers.
\newblock In {\em Proceedings of the IEEE/CVF Conference on Computer Vision and
  Pattern Recognition}, pages 8741--8750, 2021.

\bibitem{wu2021cvt}
Haiping Wu, Bin Xiao, Noel Codella, Mengchen Liu, Xiyang Dai, Lu Yuan, and Lei
  Zhang.
\newblock Cvt: Introducing convolutions to vision transformers.
\newblock In {\em Proceedings of the IEEE/CVF International Conference on
  Computer Vision}, pages 22--31, 2021.

\bibitem{wu2022tinyvit}
Kan Wu, Jinnian Zhang, Houwen Peng, Mengchen Liu, Bin Xiao, Jianlong Fu, and Lu
  Yuan.
\newblock Tinyvit: Fast pretraining distillation for small vision transformers.
\newblock {\em arXiv preprint arXiv:2207.10666}, 2022.

\bibitem{xu2019data}
Ting-Bing Xu and Cheng-Lin Liu.
\newblock Data-distortion guided self-distillation for deep neural networks.
\newblock In {\em Proceedings of the AAAI Conference on Artificial
  Intelligence}, volume~33, pages 5565--5572, 2019.

\bibitem{Xu_2021_ICCV}
Weijian Xu, Yifan Xu, Tyler Chang, and Zhuowen Tu.
\newblock Co-scale conv-attentional image transformers.
\newblock In {\em Proceedings of the IEEE/CVF International Conference on
  Computer Vision (ICCV)}, pages 9981--9990, October 2021.

\bibitem{yang2021focal}
Jianwei Yang, Chunyuan Li, Pengchuan Zhang, Xiyang Dai, Bin Xiao, Lu Yuan, and
  Jianfeng Gao.
\newblock Focal self-attention for local-global interactions in vision
  transformers, 2021.

\bibitem{yin2022avit}
Hongxu Yin, Arash Vahdat, Jose Alvarez, Arun Mallya, Jan Kautz, and Pavlo
  Molchanov.
\newblock {A}-{V}i{T}: {A}daptive tokens for efficient vision transformer.
\newblock In {\em Proceedings of the IEEE/CVF Conference on Computer Vision and
  Pattern Recognition}, 2022.

\bibitem{yu2022metaformer}
Weihao Yu, Mi Luo, Pan Zhou, Chenyang Si, Yichen Zhou, Xinchao Wang, Jiashi
  Feng, and Shuicheng Yan.
\newblock Metaformer is actually what you need for vision.
\newblock In {\em Proceedings of the IEEE/CVF Conference on Computer Vision and
  Pattern Recognition}, pages 10819--10829, 2022.

\bibitem{yun2020regularizing}
Sukmin Yun, Jongjin Park, Kimin Lee, and Jinwoo Shin.
\newblock Regularizing class-wise predictions via self-knowledge distillation.
\newblock In {\em Proceedings of the IEEE/CVF conference on computer vision and
  pattern recognition}, pages 13876--13885, 2020.

\bibitem{zhang2022minivit}
Jinnian Zhang, Houwen Peng, Kan Wu, Mengchen Liu, Bin Xiao, Jianlong Fu, and Lu
  Yuan.
\newblock Minivit: Compressing vision transformers with weight multiplexing.
\newblock In {\em Proceedings of the IEEE/CVF Conference on Computer Vision and
  Pattern Recognition}, pages 12145--12154, 2022.

\bibitem{zhang2019your}
Linfeng Zhang, Jiebo Song, Anni Gao, Jingwei Chen, Chenglong Bao, and Kaisheng
  Ma.
\newblock Be your own teacher: Improve the performance of convolutional neural
  networks via self distillation.
\newblock In {\em Proceedings of the IEEE/CVF International Conference on
  Computer Vision}, pages 3713--3722, 2019.

\bibitem{zheng2020end}
Minghang Zheng, Peng Gao, Renrui Zhang, Kunchang Li, Xiaogang Wang, Hongsheng
  Li, and Hao Dong.
\newblock End-to-end object detection with adaptive clustering transformer.
\newblock {\em arXiv preprint arXiv:2011.09315}, 2020.

\bibitem{zheng2021rethinking}
Sixiao Zheng, Jiachen Lu, Hengshuang Zhao, Xiatian Zhu, Zekun Luo, Yabiao Wang,
  Yanwei Fu, Jianfeng Feng, Tao Xiang, Philip~HS Torr, et~al.
\newblock Rethinking semantic segmentation from a sequence-to-sequence
  perspective with transformers.
\newblock In {\em Proceedings of the IEEE/CVF conference on computer vision and
  pattern recognition}, pages 6881--6890, 2021.

\bibitem{zhou2021deepvit}
Daquan Zhou, Bingyi Kang, Xiaojie Jin, Linjie Yang, Xiaochen Lian, Zihang
  Jiang, Qibin Hou, and Jiashi Feng.
\newblock Deepvit: Towards deeper vision transformer.
\newblock {\em arXiv preprint arXiv:2103.11886}, 2021.

\bibitem{zhou2002ensembling}
Zhi-Hua Zhou, Jianxin Wu, and Wei Tang.
\newblock Ensembling neural networks: many could be better than all.
\newblock {\em Artificial intelligence}, 137(1-2):239--263, 2002.

\bibitem{zhu2020deformable}
Xizhou Zhu, Weijie Su, Lewei Lu, Bin Li, Xiaogang Wang, and Jifeng Dai.
\newblock Deformable detr: Deformable transformers for end-to-end object
  detection.
\newblock In {\em International Conference on Learning Representations}, 2020.

\end{thebibliography}
% }

{\small

\bibliographystyle{IEEEtran}
}

\clearpage
\section{Appendix}

\section{Pruning residual connection in the standard form}
In our ensemble form, we cut out short paths to achieve better performance. Here, we explore whether we can obtain the same effect by pruning residual connection in the standard form. We experiment with the DeiT-S deleting the residual connection in the shallow layers and the results are shown in Table \ref{table residual}. We can see that cutting out residual connection affects the performance and convergence, which demonstrates that the success of our path pruning is not from cutting out residual connection and cannot be achieved in the standard form.

\section{Our ensemble form of hierarchical ViTs}
\label{a 8}
\begin{figure}
\includegraphics[width=1.\linewidth]{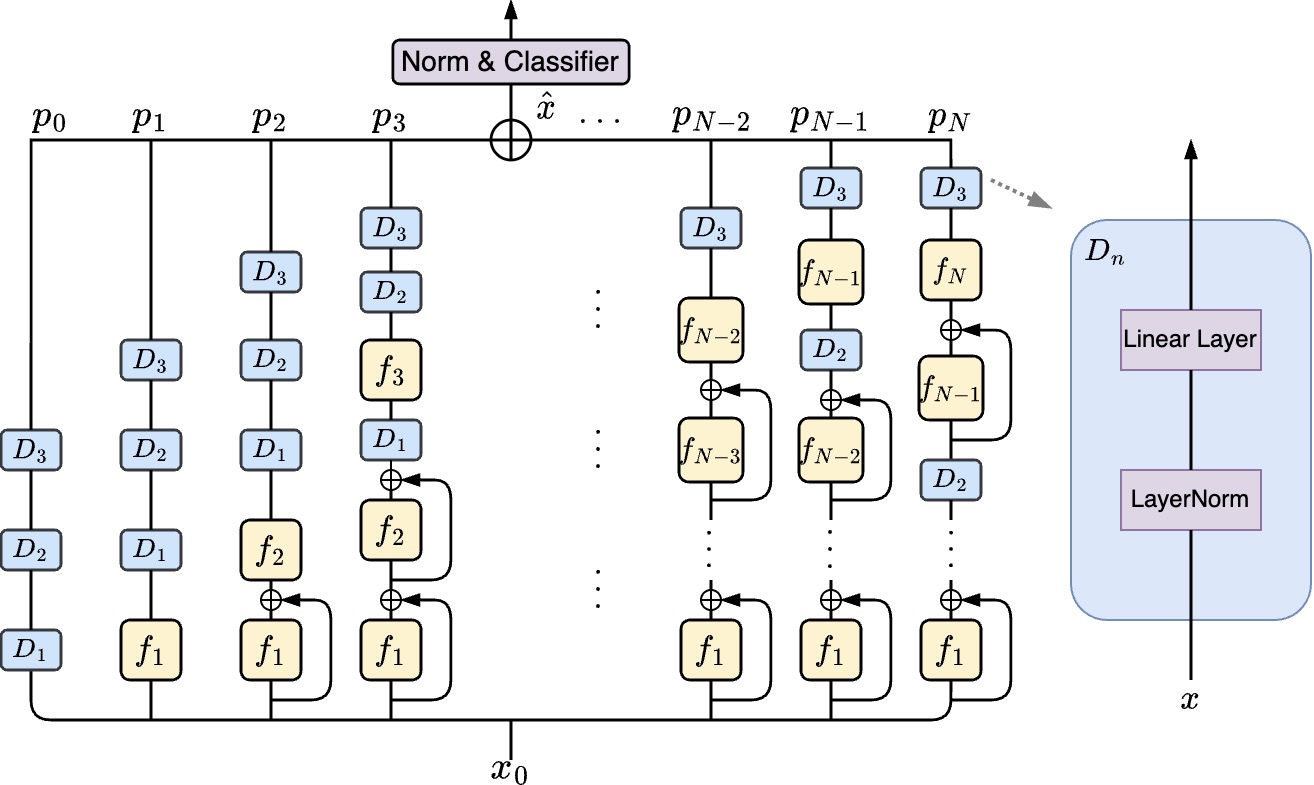}
\centering
\caption{Our ensemble form of hierarchical ViTs. $D_n$ represents the n-th downsampling layer.}
\label{hie}
\end{figure}

We visualize our ensemble form of hierarchical ViTs in Figure~\ref{hie}. 
The LayerNorm expression in our model is $E[x]/\sqrt{Var[x]}*\gamma+\beta$. In Figure~\ref{hie}, we observe that the same downsampling layer $D_n$ in different paths compute individual standard deviations, namely asynchronous standard deviation, causing different forward propagation result with standard form. 
Neglecting the influence of bias, to achieve consistent forward propagation, we need to synchronize standard deviations in different paths, namely synchronous standard deviation. 
For example, the input of $D_1$ in $p_0$, $p_1$, $p_2$, and $p_3$ are different, leading to different standard deviations. The input of $D_1$ in $p_3$ is the same as the standard form. Therefore, if we want to achieve the same forward propagation, we can synchronize all the standard deviations of $D_1$ with the standard deviation in $p_3$.
However, we find that using either asynchronous or synchronous standard deviation yields similar performance when we train them from scratch.

\section{Self-distillation in the standard form}
We apply our self-distillation method in the standard form to make low-level feature maps mimic high-level feature maps in Table~\ref{table dis} and find out that it is difficult to work. The models suffer from an accuracy drop or divergence. We try to explain this issue from an ensemble perspective. 

Assuming that we select $x_t$ and $x_s$ ($t > s$) which are the output of any intermediate transformers $T_t$ and $T_s$ as the teacher and the student, respectively. There are $t-s$ transformers between $x_t$ and $x_s$. According to the Eq. \textcolor{red}{5}, we can find a function $\mathcal{F}$ and denote the $x_t$ as
\begin{equation}
    x_{t} = x_s + \mathcal{F}(x_s),
\end{equation}
where a student component is in the teacher feature map. Then we force the $x_s$ to mimic the $x_s + \mathcal{F}(x_s)$, \ie, $x_t$. The model may be optimized to an unexpected direction by the KD loss, such as enlarging the weight of $x_s$ in $x_t$ and decreasing the $\mathcal{F}(x_s)$ to $0$. When we use $l_2$ loss as the KD loss, the effect is most obvious where the model diverges directly. Therefore, we speculate that the inherent ensemble property of ViTs limits the application of self-distillation in the standard form. In contrast, our ensemble view avoids this issue. Our ensemble form decouples the linear combination and the paths do not contain the linear components of previous paths. 

\begin{table}[t]
\small
\begin{center}
\begin{tabular}{c|c|c}
            \toprule
            Model & No. of layers w/o shortcut & Accuracy    \\
            \midrule
            DeiT-S & 0 (Baseline) & 79.8  \\
            DeiT-S & 1 & 77.7  \\
            DeiT-S & 2 & Loss NAN  \\
            \bottomrule
        \end{tabular}
\end{center}
\caption{Pruning the residual connection in the shallow transformer layers.}
\label{table residual}
\end{table}

\begin{table}[t]
\small
\begin{center}
\begin{tabular}{c|c|c}
            \toprule
            Model & KD Loss & Accuracy    \\
            \midrule
            DeiT-S & - & 79.8 (Baseline)  \\
            DeiT-S & $l_2$ Loss & Loss NAN  \\
            DeiT-S & KL Loss & 79.6  \\
            \bottomrule
        \end{tabular}
\end{center}
\caption{Applying our self-distillation method to distill feature maps in the standard ViT form.\vspace{-3mm}}
\label{table dis}
\end{table}

\begin{table}[t]
\small
\begin{center}
\begin{tabular}{c|c|c|c}
            \toprule
            SA path & FFN path  & ES & Accuracy (\%)    \\
            \midrule
            $p_8$ - $p_{12}$ & $p_1$ - $p_{12}$ &  & 80.0 \\
            $p_8$ - $p_{12}$ & $p_3$ - $p_{12}$ &  &  80.1 \\
            $p_1$ - $p_{12}$ & $p_1$ - $p_{12}$ & \checkmark & 80.3 \\
            \bottomrule
        \end{tabular}
\end{center}
\caption{Applying path pruning and EnsembleScale to DeiT-S with $2N+1$ paths. ES is short for Ensemble. Note that $x_0$ path is not contained in any experments.}
\label{table 2n+1}
\end{table}

\section{Path combination for 2N+1 paths}
In Eq. \textcolor{red}{5}, we combine the MHSA and FFN paths into an $f$ path and obtain $N+1$ paths in a ViT, where $N$ is the number of transformer layers. If we do not combine them, we will get $2N+1$ paths. We conduct experiments to explore the path combination for $2N+1$ paths. According to the previous works [\textcolor{green}{2}, \textcolor{green}{29}, \textcolor{green}{32}, \textcolor{green}{39}], self-attention and FFN can be regarded as low-pass filters and  high-pass filters separately. Therefore, we prefer to save more FFN paths and cut out self-attention paths. The results are presented in Table~\ref{table 2n+1}. In our experiments, we do not discover that splitting self-attention and FFN paths brings more improvement than combining them but EnsembleScale costs double parameter number.

% \section{Experiment details of efficient dynamic ViTs.}
% We add the $ES_1$ and $ES_2$ on a pre-trained DeiT-S and apply the classifier for the first 7 paths besides all the paths to build our dynamic DeiT-S. We fine-tune the parameters of $ES_1$, $ES_2$, and the classifier for 200 epochs. The objective loss function is the sum of two classification loss terms: one for the first 7 paths and the other for all paths. During inference, we set the confidence score threshold to 0.6. If an image is given, it will be predicted by the first 7 paths. The inference process stops once the prediction score exceeds 0.6; otherwise, we use the final prediction as the result.

% The results are presented in Table~\ref{table dynamic}. After adding $ES_1$ and $ES_2$ and finetuning, the accuracy of DeiT-S increases to 80.0\% when evaluating the validation set using all the paths. Subsequently, we apply our dynamic ViT to this network, resulting in a 25\% reduction in FLOPs and an accuracy of 79.8\% which is the same as the accuracy of the original DeiT-S. The number of data processed by different stages of our dynamic ViT is shown in Table~\ref{table count}. It is observed that approximately 51.6\% ``easy" image are processed using the first 7 paths with an accuracy of 93.1\%, while the remaining 48.4\% ``hard" data require  the whole network. 
% Our dynamic ViT is the simplest implementation. We believe that more effective and efficient dynamic ViT can be achieved based on our ensemble form.

\section{The demo code of our ensemble form}
The demo code of our ensemble form is summarized in Algorithm~\ref{alg}. We only require a few modifications in the code of the standard form, demonstrating our ensemble form is implementation- and deployment-friendly.

% \begin{table}[t]
% \small
% \begin{center}
% \begin{tabular}{c|c|c}
%             \toprule
%             Model & FLOPs(G) & Accuracy(\%)    \\
%             \midrule
%             DeiT-S & 4.58 & 79.8  \\
%             DeiT-S (finetuning) & 4.58 & 80.0  \\
%             Dynamic DeiT-S & 3.42(-25\%) & 79.8  \\
%             \bottomrule
%         \end{tabular}
% \end{center}
% \caption{We compare the our dynamic DeiT-S with the original DeiT-S model, reporting their respective top-1 accuracy and
% FLOPs.}
% \label{table dynamic}
% \end{table}

% \begin{table}[t]
% \small
% \begin{center}
% \begin{tabular}{c|c|c}
%             \toprule
%              & First 7 paths & All the paths    \\
%             \midrule
%             No. of images & 25824 & 24176  \\
%             Accuracy(\%) & 93.1 & 65.5  \\
%             \bottomrule
%         \end{tabular}
% \end{center}
% \caption{The data distribution of our dynamic ViT.}
% \label{table count}
% \end{table}

\begin{algorithm} 
\small
	\caption{Demo code of our ensemble form (PyTorch-like)} 
	\label{alg} 
	\begin{algorithmic}
        \STATE \textcolor[RGB]{81,136,79}{\# N: the number of transformer layers}
        \STATE \textcolor[RGB]{81,136,79}{\# self\_attention: the function of self attention}
        \STATE \textcolor[RGB]{81,136,79}{\# ffn: the function of FFN}
        \STATE \textcolor[RGB]{81,136,79}{\# patch\_embedding: the function of patch embedding}
        \STATE
		 \STATE class Block: \\
                 \STATE \quad def forward (input): \\
                 \STATE \quad \quad sa\_path = self\_attention(norm(input)) \\
                 \STATE \quad \quad ffn\_path = ffn(norm(input + sa\_path)) \\
                 \STATE \quad \quad return input + sa\_path + ffn\_path, sa\_path + ffn\_path \\
        \STATE 
        \STATE class ViT: \\
        \STATE \quad def init()
        \STATE \quad \quad blocks = [Block() for i in range(N)]
        \STATE
        \STATE \quad def forward(input):
        \STATE \quad \quad x = patch\_embedding(input)
        \STATE \quad \quad paths = [x]
        \STATE \quad \quad for i in range(N):
        \STATE \quad \quad \quad x, f = blocks[i](x)
        \STATE \quad \quad \quad paths.append(f)
        \STATE \quad \quad return sum(paths)
	\end{algorithmic} 
\end{algorithm}

\end{document}